\documentclass[twocolumn]{IEEEtran}

\usepackage[pdftex]{graphicx} 
\usepackage{amsmath} 
\usepackage{algorithmic}
\usepackage{array} 
\usepackage{placeins}
\usepackage[caption=false,font=footnotesize]{subfig} 
\usepackage{dblfloatfix}
\usepackage{url} 

\usepackage{microtype}
\usepackage[utf8]{inputenc} 
\usepackage[T1]{fontenc}    
\usepackage{paralist} 
\usepackage{marvosym} 
\usepackage{bm} 
\usepackage{bbm} 
\usepackage{amssymb}   
\usepackage{mathrsfs} 
\usepackage{tabularray} 
\UseTblrLibrary{booktabs}       
\usepackage{tabularx} 
\usepackage[flushleft]{threeparttable} 
\usepackage{colortbl} 
\usepackage{makecell} 
\usepackage{multirow} 
\usepackage{nicefrac} 
\usepackage[dvipsnames]{xcolor} 
\usepackage{lettrine} 
\usepackage{amsmath}
\usepackage{mathtools}  
\usepackage{soul} 
\setul{0.5ex}{0.3ex}

\definecolor{citecolor}{RGB}{34,139,34}
\usepackage[pagebackref=true,breaklinks=true,letterpaper=true,colorlinks,
    citecolor=citecolor,bookmarks=false]{hyperref}
\usepackage{cleveref}  
\crefformat{figure}{Fig.~#2#1#3}
\crefformat{equation}{Eq.~(#2#1#3)}
\crefformat{table}{Table~#2#1#3}
\crefformat{section}{Section~#2#1#3}
\crefformat{algorithm}{Algorithm~#2#1#3}
\crefformat{appendix}{Appendix #2#1#3}
\crefformat{subsection}{subsection~#2#1#3}
\usepackage{cellspace}
\usepackage[numbers,sort&compress]{natbib} 

\usepackage{xspace} 

\usepackage{pifont} 

\definecolor{modifiedorange}{RGB}{255,140,0} 
\definecolor{modifiedblue}{RGB}{64,120,192}

\definecolor{Gray}{rgb}{0.9,0.9,0.9}
\definecolor{LightCyan}{rgb}{0.88,1,1}
\newcolumntype{a}{>{\columncolor{Gray}}c}
\newcolumntype{b}{>{\columncolor{white}}c}

\begin{document}

\setlength{\abovedisplayskip}{.5\baselineskip} 
\setlength{\belowdisplayskip}{.5\baselineskip} 

\title{DRPCA-Net: Make Robust PCA Great Again for Infrared Small Target Detection}


\author{
  Zihao~Xiong,
  Fei~Zhou,
  Fengyi~Wu,
  Shuai~Yuan,
  Maixia~Fu,
  Zhenming~Peng,
  Jian~Yang,
  Yimian~Dai
  \thanks{
    This work was supported by
      the National Natural Science Foundation of China (62303165, 
      62301261, 
      62361166670, U24A20330), 
      the Open Project of Key Laboratory of Grain Information Processing and Control, Ministry of Education, Henan University of Technology, under Grant KFJJ2023014,
    \emph{(Corresponding author:
     Maixia Fu and Yimian Dai).}
    }

  \thanks{Zihao Xiong, Fei Zhou, and Maixia Fu are with Key Laboratory of Grain Information Processing and Control (Henan University of Technology), Ministry of Education; Henan Key Laboratory of Grain Photoelectric Detection and Control, Henan University of Technology, Zhengzhou, China.
  (e-mail: 
  \href{mailto:zihhao.syong@gmail.com}{zihhao.syong@gmail.com};
  \href{mailto:hellozf1990@163.com}{hellozf1990@163.com};
  \href{fumaixia@126.com}{fumaixia@126.com}).
  }

  \thanks{Fengyi Wu and Zhenming Peng are with the School of Information and Communication Engineering and the Laboratory of Imaging Detection and Intelligent Perception, University of Electronic Science and Technology of China, Chengdu, China. (e-mail: \href{mailto:wufengyi98@163.com}{wufengyi98@163.com}; \href{mailto:zmpeng@uestc.edu.cn}{zmpeng@uestc.edu.cn}).}

  \thanks{Shuai~Yuan is with the School of Optoelectronic Engineering, Xidian University, Xi'an 710071, China. (email: \href{mailto:yuansy@stu.xidian.edu.cn}{yuansy@stu.xidian.edu.cn}).}
  \thanks{Jian Yang and Yimian Dai are with PCA Lab, VCIP, College of Computer Science, Nankai University. Y. Dai also holds a position at the NKIARI, Shenzhen Futian. (e-mail:
  \href{mailto:csjyang@nankai.edu.cn}{csjyang@nankai.edu.cn};
  \href{mailto:yimian.dai@gmail.com}{yimian.dai@gmail.com}).
  }

}

\maketitle


\begin{abstract}

Infrared small target detection plays a vital role in remote sensing, industrial monitoring, and various civilian applications.
Despite recent progress powered by deep learning, many end-to-end convolutional models tend to pursue performance by stacking increasingly complex architectures, often at the expense of interpretability, parameter efficiency, and generalization.
These models typically overlook the intrinsic sparsity prior of infrared small targets--an essential cue that can be explicitly modeled for both performance and efficiency gains.
To address this, we revisit the model-based paradigm of Robust Principal Component Analysis (RPCA) and propose Dynamic RPCA Network (DRPCA-Net), a novel deep unfolding network that integrates the sparsity-aware prior into a learnable architecture.
Unlike conventional deep unfolding methods that rely on static, globally learned parameters, DRPCA-Net introduces a dynamic unfolding mechanism via a lightweight hypernetwork.
This design enables the model to adaptively generate iteration-wise parameters conditioned on the input scene, thereby enhancing its robustness and generalization across diverse backgrounds.
Furthermore, we design a Dynamic Residual Group (DRG) module to better capture contextual variations within the background, leading to more accurate low-rank estimation and improved separation of small targets.
Extensive experiments on multiple public infrared datasets demonstrate that DRPCA-Net significantly outperforms existing state-of-the-art methods in detection accuracy.
Code is available at \href{https://github.com/GrokCV/DRPCA-Net}{https://github.com/GrokCV/DRPCA-Net}.

\end{abstract}

\vspace{-1\baselineskip}

\section{Introduction} \label{sec:introduction}

\IEEEPARstart{O}{wing} to thermal imaging's unique capacity for passive, illumination-free, and long-range detection, infrared small target detection (IRSTD) is a cornerstone technology for various remote sensing applications, including environmental monitoring, disaster rescue, and industrial inspection~\cite{TGRS2021ALCNet}.
Nevertheless, the primary challenge in IRSTD stems from the need to accurately discern distant objects, which appear as minute, indistinct points within infrared imagery due to their substantial distance from the sensor~\cite{TGRS2023Dim2Clear}.
Compounding this difficulty, such targets typically occupy only a handful of pixels, possess low signal-to-noise ratios, and lack identifiable texture or shape characteristics, thereby rendering IRSTD a formidable and enduring research problem~\cite{TGRS20243DST}.

\begin{figure*}[htbp]
\centering
\includegraphics[scale=0.45]{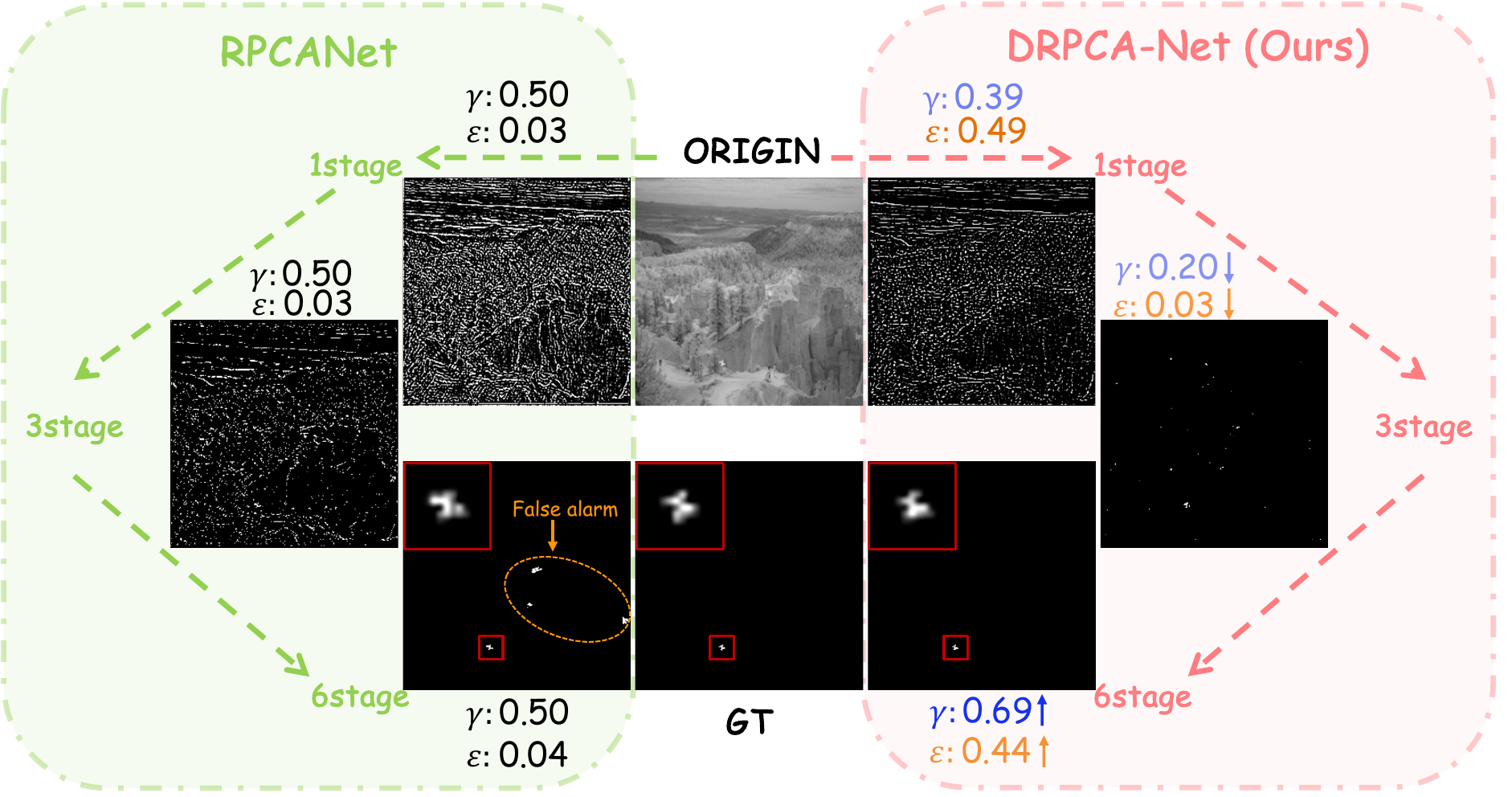}
\caption{Visual comparison between conventional static unfolding (RPCANet) and our proposed dynamic unfolding (DRPCA-Net). The visualizations at different stages (1, 3, and 6) demonstrate that the \textbf{dynamic parameter generation mechanism helps to suppress background interference effectively}.} 
\label{fig:motivation}
\end{figure*}

\subsection{Prior Works on Infrared Small Target Detection}

Existing IRSTD methods can be broadly classified into model-driven and data-driven approaches. Model-driven methods are typically based on handcrafted priors, such as local contrast~\cite{TGRS2014LCM,LGRS2019TLLCM}, sparsity~\cite{TGRS2025IWS}, and background low-rankness~\cite{JSTARS2017RIPT}. These techniques often employ simple features, e.g., grayscale intensity, gradient, or entropy~\cite{GRSL2019RDLCM}, and perform well in simpler scenes with consistent backgrounds.
However, their reliance on fixed assumptions and manually tuned hyperparameters significantly limits their adaptability in complex environments. 
As the scene becomes more heterogeneous and the signal-to-noise ratio decreases, the effectiveness of such methods quickly deteriorates, revealing poor generalization across varying imaging conditions~\cite{JSTARS2024ISPANet,JSTARS2025MDIGCNet}.

Recently, the advent of deep learning has fully revolutionized the field of infrared small target detection, with data-driven methods achieving remarkable success by harnessing the robust feature extraction and representation capabilities of neural networks.
Notable examples include architectures that leverage attention mechanisms~\cite{TGRS2021ACM}, generative adversarial networks (GANs)~\cite{TGRS2022GAN}, and multi-scale feature fusion~\cite{TIP2023DNANet}, all of which have demonstrated significant improvements in detection accuracy.
However, the pursuit of marginal accuracy gains has led these models toward increasingly complex designs coupled with high-resolution feature maps, resulting in substantial computational burdens.
This becomes particularly problematic in real-world IRSTD scenarios, where real-time processing on resource-constrained platforms (e.g., airborne or embedded systems) is often required.

We argue that the fundamental limitation of current end-to-end approaches stems from treating IRSTD as a generic object detection problem, \textbf{necessitating increasingly complex architectures to compensate for the absence of domain-specific priors}.
These methods overlook a critical characteristic of infrared small targets, namely their inherent sparsity within the spatial domain, which could significantly enhance detection performance while reducing model complexity if properly leveraged.
Recognizing this gap, recent research has begun exploring the integration of task-specific priors into deep learning frameworks.
Dai \textit{et al.}~\cite{TGRS2021ALCNet} pioneered this direction by embedding local saliency priors into a bottom-up attention module, guiding the network to focus on regions with high local contrast. Lin \textit{et al.}~\cite{TIP2024Learning} proposed a contrast-shape representation learning approach that utilizes central difference convolutions and large-kernel operations to extract shape-preserving features while maintaining sensitivity to low-contrast targets.
Recently, Wu \textit{et al.}~\cite{WACV2024RPCANet} and Zhou \textit{et al.}~\cite{TGRS2023DeepLSPNet} have introduced deep unfolding strategies to transform traditional low-rank sparse decomposition models into hierarchical and learnable deep frameworks, which can leverage the advantages of both global priors from low-rank sparsity and data-driven approaches.
While promising, these attempts to bridge domain-specific priors with deep architectures still reveal critical limitations in real-world IRSTD scenarios:
\begin{enumerate}
    \item \textbf{Prior Under-Utilization:} 
    Current contrast-enhanced networks implicitly guide target localization through local intensity variations but fail to explicitly encode the global sparsity of infrared small targets and the low-rankness of backgrounds. This leads to ambiguous discrimination between true sparse targets and high-contrast background noise, resulting in persistent false alarms. 
    \item \textbf{Context-Blind Prior Modeling:} 
    Existing unfolding networks employ fixed convolutional kernels for background estimation, assuming uniform spatial correlations across scenes. As shown in Fig.~\ref{fig:motivation}, this rigid approach conflicts with the heterogeneous nature of infrared backgrounds, causing either over-smoothed structures or residual clutter.
\end{enumerate}


\subsection{Motivation}

To address the above limitations, we revisit the classical Robust Principal Component Analysis (RPCA) formulation from a new perspective.
Specifically, we ask: \textit{Can we build a learnable yet prior-aware architecture that dynamically adapts its optimization trajectory to the input scene, while preserving the core low-rank/sparse decomposition principles of RPCA?}
Our answer is affirmative. Drawing inspiration from the concept of hypernetworks, we introduce a dynamic unfolding paradigm that enables adaptive, iteration-wise parameter generation conditioned on the input. 
This \textbf{context-conditioned prior unfolding} mechanism allows the network to retain the mathematical clarity of RPCA while gaining the flexibility of deep learning.

Building on this idea, we propose the Dynamic Robust Principal Component Analysis Network (DRPCA-Net), a novel deep unfolding architecture that we refer to as a dynamic unfolding framework. Unlike conventional designs, DRPCA-Net decouples the parameterization of each iteration step from static training and instead generates it dynamically via a lightweight hypernetwork.
We introduce a set of core parameters, such as update coefficients and balancing factors, which govern the iterative decomposition process. In contrast to traditional deep unfolding methods that learn these parameters globally and keep them fixed across all inputs, DRPCA-Net predicts them in a scene-aware manner. This dynamic adjustment enables the model to adapt its inference trajectory to the specific characteristics of each input, enhancing flexibility and generalization across diverse scenarios.

Beyond the optimization parameters, we also propose the Dynamic Residual Group (DRG) module to model the image reconstruction in a context-aware manner.
Traditional image modeling modules rely on fixed convolutional kernels, which are insufficient for capturing the heterogeneous spatial statistics of infrared scenes.
Our DRG module integrates residual learning with dynamic spatial attention mechanisms, which replaces static convolutional operations with content-adaptive filters to better capture spatially varying background patterns and structural dependencies.

Extensive experiments on multiple public infrared small target detection datasets validate the efficacy of our approach, in which DRPCA-Net consistently outperforms state-of-the-art methods in both detection accuracy and computational efficiency.
It demonstrates that our input-conditioned parameterization paradigm significantly enhances the model's robustness across varying IRSTD conditions.

The main contributions of this work are summarized as follows:
\begin{enumerate}
    \item \textbf{Dynamic Prior-Aware Unfolding:} We propose a novel deep unfolding framework that integrates RPCA priors with a dynamic parameter generation mechanism via hypernetworks, enabling input-conditioned optimization.
    \item \textbf{Dynamic Residual Group:} We design a DRG module that combines residual connections with a dynamic attention mechanism to improve image modeling and enhance target-background separation.
    \item \textbf{State-of-the-Art Performance:} DRPCA-Net achieves new benchmarks on multiple IRSTD datasets in terms of accuracy and robustness, demonstrating its practical viability.
\end{enumerate}


\section{Related Work} \label{sec:related}

\subsection{Infrared Small Target Detection} \label{subsec:IRSTD}

The evolution of infrared small target detection reflects a paradigm shift from low-level image processing to high-level semantic understanding. Early approaches relied on heuristic feature engineering based on grayscale variance and local contrast, treating detection as an anomaly identification problem. With the advent of deep learning, the field has transitioned toward data-driven representation learning, where hierarchical features are automatically discovered through neural architectures. This progression mirrors broader trends in computer vision, yet retains unique challenges due to the extreme sparsity and weak signature of infrared targets~\cite{TGRS2024GCINet}.

Among model-driven techniques, low-rank and sparse decomposition (LRSD) methods have emerged as a principled framework for IRSTD by explicitly modeling the structural priors of infrared imagery: sparse targets embedded in low-rank backgrounds. Representative works such as the Infrared Patch-Image (IPI) model~\cite{TIP2013IPI} formulate the detection task as a constrained optimization problem, separating targets from backgrounds via global low-rank approximation and spatially sparse regularization. Subsequent efforts introduced non-convex relaxations~\cite{Jinfrared2017NIPPS,Jinfrared2016WIPI}, tensor-based decompositions~\cite{JSTARS2017RIPT,TAES2023CMPG}, and spatial-structural regularizations~\cite{TGRS2023nonconvextensorTD,TIP2020TNLRS} to better capture the intrinsic global and local structure of infrared scenes. These methods achieve interpretable decompositions and remain competitive in scenarios with limited data.

Despite their mathematical elegance, model-driven approaches suffer from several intrinsic limitations:
\begin{enumerate}
    \item \textbf{Idealized Priors:}
    The assumption that targets are strictly sparse and backgrounds are perfectly low-rank is often violated in real-world scenarios. Many background textures, such as small clouds, sea clutter, and terrain features, exhibit sparse-like behavior, leading to a high false alarm rate when relying solely on low-rank/sparse separation.
    \item \textbf{Hyperparameter Sensitivity:} 
    Most LRSD models require careful tuning of multiple hyperparameters, such as regularization weights and rank thresholds, which are typically fixed across scenes. 
    This rigidity hinders their ability to generalize to varying target sizes, contrast levels, and noise distributions, making them less applicable to dynamic and heterogeneous infrared environments.
    \item \textbf{Computational Inefficiency:}
    The matrix decomposition in conventional RPCA-based methods imposes prohibitive computational burdens.
    Specifically, the singular value decomposition (SVD) operation required for low-rank approximation scales cubically with matrix dimension $(O(n^3))$, becoming intractable for high-resolution imagery.
\end{enumerate}

The emergence of deep learning has revolutionized the IRSTD field by addressing these limitations through data-driven feature learning. Wang \textit{et al.}~\cite{ICCV2019MDvsFAcGAN} introduced conditional generative adversarial networks to balance detection sensitivity and specificity, while Dai \textit{et al.}~\cite{TGRS2021ACM} established a standardized benchmark dataset and proposed asymmetric contextual modeling for enhanced feature representation. 
Subsequent research has explored various architectures, including specialized convolutional networks~\cite{TAES2022LPNet}, vision transformers~\cite{TGRS2023MTUNet, TGRS2024SCTransNet}, state-space models~\cite{TGRS2024MiMISTD}, U-shaped architectures~\cite{TGRS2024U2AMFPNet}, and segment-anything adaptations~\cite{ECCV2024IRSAM}. 
These approaches leverage the representational capacity of deep networks to automatically extract discriminative features without explicit prior formulations. However, this shift toward increasingly complex architectures introduces new challenges: these models typically demand substantial training data to avoid overfitting, a fundamental issue in the data-scarce IRSTD domain. Besides, their computational load often exceeds the constraints of real-time detection systems deployed on resource-limited platforms.

To address these limitations, our DRPCA-Net differs from existing approaches with the context-aware unfolding paradigm, which fundamentally transforms the rigid decomposition framework into a flexible architecture that maintains mathematical interpretability while achieving state-of-the-art performance in challenging IRSTD environments.
\begin{enumerate}
    \item We explicitly encode the sparsity and low-rank priors within a learnable architecture, enabling our model to achieve superior detection performance with significantly fewer parameters compared to conventional deep learning methods that lack such structured inductive biases. 
    \item We introduce a dynamic parameter generation mechanism that produces iteration-specific and input-dependent parameters for the unfolding process, enabling adaptive decomposition across diverse operational scenarios without manual recalibration. 
\end{enumerate}

\subsection{Deep Unfolding Networks} \label{subsec:DU}

Deep unfolding networks (DUNs) represent a principled paradigm that integrates model-based optimization and data-driven learning. Instead of treating neural networks as black-box function approximators, DUNs unfold classical iterative algorithms into layer-wise neural architectures, enabling end-to-end training while preserving algorithmic interpretability. This approach not only embeds domain knowledge into the network structure but also enhances generalization under limited training data--a particularly desirable property for tasks constrained by data scarcity or requiring strong inductive priors.

The origins of DUNs trace back to the Learned Iterative Shrinkage-Thresholding Algorithm (LISTA)~\cite{ICLM2010LISTA}, which parameterized the ISTA process via learnable matrices. Later, ISTA-Net~\cite{TNNLS2023ISTA} introduced convolutional variants to better capture spatial dependencies in images. ISTA-Net+~\cite{ICME2021ISTANET++} and ISTA-Net++~\cite{ICLR2017understanding} further improved feature propagation and representational capacity through the use of long-range skip connections and channel modulation (CM) structures. Meanwhile, FISTA-Net~\cite{NIPS2016Maximal} extended the unfolding of the accelerated FISTA algorithm, allowing adaptive learning of step sizes and thresholds. These advances collectively demonstrate how unfolding strategies can enhance convergence speed, representation learning, and robustness, especially in ill-posed inverse problems.

Beyond sparse coding and compressed sensing, DUNs have been successfully extended to a variety of imaging and vision tasks. For instance, ADMM-CSNet~\cite{TPAMI2020ADMMCSNet} leverages the unfolded ADMM framework for MRI reconstruction, while AMP-inspired unfolding has been employed for denoising~\cite{TIP2020AMPNet} and artifact reduction in CT imaging~\cite{IJCAI2022Proceedings}. In the domain of hyperspectral and HDR imaging, attention-augmented unfolding networks have shown strong performance by capturing both global structure and local variations~\cite{TPAMI2024AGTC}. Furthermore, contrastive learning has been integrated into DUNs to improve feature discrimination in image deraining tasks~\cite{TNNLS2024CUDEN}. These works collectively validate the versatility and extensibility of the unfolding framework across domains.

Despite such promising advances, the application of DUNs to the field of IRSTD remains rarely explored. Existing efforts, such as RPCANet~\cite{WACV2024RPCANet}, have attempted to integrate RPCA-inspired priors into deep unfolding schemes. However, the challenges of fixed parameterization and context-agnostic modeling still persist.
Our work addresses these limitations by introducing scene-conditioned parameter adaptation within the unfolding framework. Unlike conventional approaches that learn fixed parameters during training, DRPCA-Net employs a lightweight hypernetwork that generates iteration-specific parameters conditioned on input characteristics.


\section{Method} \label{sec:method}
\begin{figure*}[htbp]
\centering
\includegraphics[width=\textwidth]{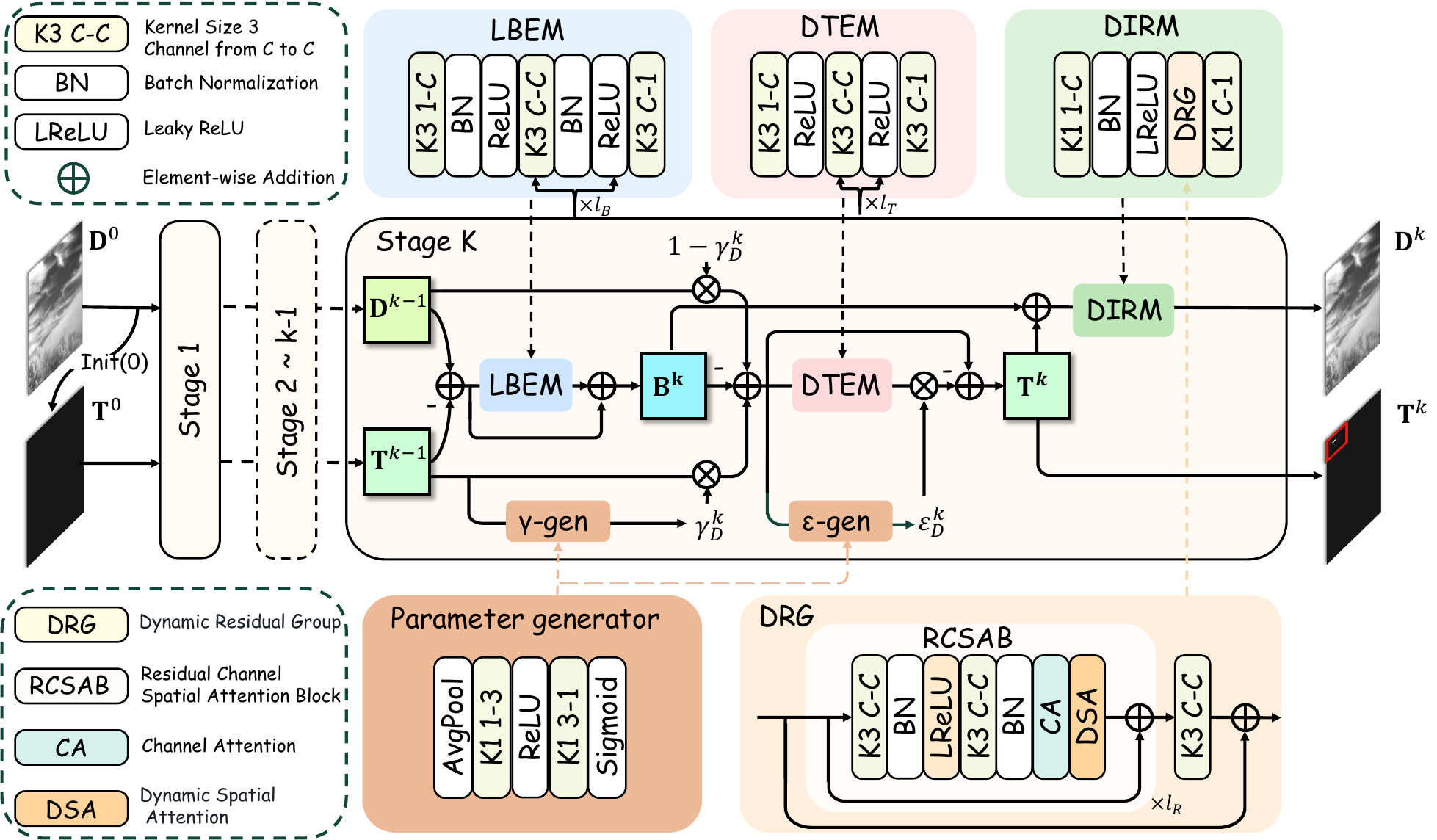}
\caption{Overall structure of DRPCA-Net. The network is composed of K stages, and the structure of each stage is the same. The detailed structure of the K-th stage is presented, which includes: latent background encoder module (LBEM), dynamic target extraction module (DTEM), dynamic image reconstruction module (DIRM), and a parameter generator. Among them, dynamic spatial attention (DSA) belongs to DIRM.}
\label{fig:drpcanet}
\end{figure*}

\subsection{Overall Architecture of DRPCA-Net}
\label{subsec:overall_architecture}
DRPCA-Net establishes a principled synergy between the model-based interpretability of RPCA and the adaptive representation power of dynamic deep networks. By conceptualizing the iterative optimization process inherent to RPCA through the lens of deep unfolding, DRPCA-Net materializes as a learnable, stage-wise architecture specifically tailored for infrared small target detection. Crucially, it introduces a dynamic parameterization mechanism, enabling adaptive inference conditioned on the input scene statistics, thereby transcending the limitations of conventional unfolding networks with fixed parameters.

\subsubsection{Revisiting Robust Principal Component Analysis}

RPCA serves as a cornerstone for modeling infrared scenes, postulating that an observed infrared image $\mathbf{D} \in \mathbb{R}^{H \times W}$ can be intrinsically decomposed into two constituent components:
\begin{equation}
\mathbf{D} = \mathbf{B} + \mathbf{T}
\label{eq:rpca_decomp}
\end{equation}
where $\mathbf{B} \in \mathbb{R}^{H \times W}$ represents the low-rank background, encapsulating spatially correlated clutter and structures, and $\mathbf{T} \in \mathbb{R}^{H \times W}$ denotes the sparse foreground, primarily capturing the localized signatures of small targets alongside potential impulsive noise. This decomposition elegantly reflects the inherent priors of typical infrared imagery: backgrounds exhibit significant spatial redundancy, while targets manifest as sparse, isolated entities.

To render the decomposition amenable to optimization within a learning framework, we consider a relaxed unconstrained objective with penalized regularization terms:
\begin{equation}
\min_{\mathbf{B}, \mathbf{T}} \mathcal{L}(\mathbf{B}, \mathbf{T}) = \mathcal{R}(\mathbf{B}) + \lambda \mathcal{S}(\mathbf{T}) + \frac{\mu}{2} \left\| \mathbf{D} - \mathbf{B} - \mathbf{T} \right\|_F^2
\label{eq:rpca_loss}
\end{equation}
where $\mathcal{R}(\cdot)$ enforces the low-rank or smoothness prior on the background $\mathbf{B}$, while $\mathcal{S}(\cdot)$ promotes the sparsity prior on the target component $\mathbf{T}$. The hyperparameter $\lambda > 0$ balances the influence of the target prior, and $\mu > 0$ controls the weight of the data fidelity term, ensuring the reconstructed components adhere to the original observation $\mathbf{D}$. This objective function is typically minimized through iterative algorithms, alternating updates for $\mathbf{B}$ and $\mathbf{T}$.

\subsubsection{Deep Unfolding RPCA into DRPCA-Net}

Inspired by the deep unfolding paradigm, we reinterpret the iterative solution trajectory of Eq. \eqref{eq:rpca_loss} as a deep neural network composed of $K$ cascaded stages. Each stage $k \in \{1, \dots, K\}$ refines the estimates of the background and target components based on the outputs of the preceding stage. Let $\mathbf{B}^k$, $\mathbf{T}^k$, and $\mathbf{D}^k$ denote the estimated background, target, and reconstructed image at the $k$-th stage, respectively. The process is initialized with the input image $\mathbf{X}$, setting $\mathbf{D}^0 = \mathbf{X}$ and the initial target estimate $\mathbf{T}^0 = \mathbf{0}$. Within each stage $k$, the updates are performed by three specialized modules, corresponding to the optimization subproblems for $\mathbf{B}$, $\mathbf{T}$, and ensuring data consistency.

\paragraph{Latent Background Encoder Module (LBEM)}
In classical iterative RPCA solvers, the background update step aims to minimize $\mathcal{R}(\mathbf{B}) + \frac{\mu}{2} \|\mathbf{D} - \mathbf{B} - \mathbf{T}^{k-1}\|_F^2$ with respect to $\mathbf{B}$. This often involves applying a proximal operator associated with the chosen low-rank prior (e.g., singular value thresholding for the nuclear norm). Instead of relying on explicit, often computationally demanding, matrix operations, DRPCA-Net employs a learnable LBEM. The LBEM implicitly captures the background characteristics by learning a corrective mapping $\mathcal{G}_\theta^k$:
\begin{equation}
    \mathbf{B}^{k} = (\mathbf{D}^{k-1} - \mathbf{T}^{k-1}) + \mathcal{G}_\theta^k(\mathbf{D}^{k-1} - \mathbf{T}^{k-1})
    \label{eq:lbem_update}
\end{equation}
where $\mathcal{G}_\theta^k$ is realized as a lightweight convolutional network with stage-specific parameters $\theta^k$, whose architecture corresponds to the LBEM illustrated in Fig.~\ref{fig:drpcanet}, with the number of residual blocks $l_B$ is empirically set to 6 in our experiments.
It operates on the current residual $(\mathbf{D}^{k-1} - \mathbf{T}^{k-1})$, predicting the refinement needed to isolate the underlying low-rank background structure.

\paragraph{Dynamic Target Extraction Module (DTEM)}
The target update step traditionally minimizes $\lambda \mathcal{S}(\mathbf{T}) + \frac{\mu}{2} \|\mathbf{D} - \mathbf{B}^{k} - \mathbf{T}\|_F^2$ with respect to $\mathbf{T}$, often solved via proximal gradient descent. For instance, using an $l_1$ norm for $\mathcal{S}(\cdot)$ leads to iterative soft-thresholding. Prior unfolding works (e.g., based on ISTA or FISTA variants, such as the formulation adapted in~\cite{WACV2024RPCANet}) established update rules resembling momentum-based gradient steps:
\begin{equation}
\mathbf{T}^{k} \approx \gamma \mathbf{T}^{k-1} + (1 - \gamma )(\mathbf{D}^{k-1} - \mathbf{B}^{k}) - \varepsilon \cdot \nabla_{\mathbf{T}}\mathcal{S}(\mathbf{T}_{\text{interim}})
\label{eq:target_update_classical}
\end{equation}
where $\mathbf{T}_{\text{interim}}$ is an intermediate estimate, while define $\gamma = \frac{\lambda L}{\lambda L + \mu}$ and $\varepsilon = \frac{\lambda}{\lambda L + \mu}$, The detailed derivation process can be found in Ref.~\cite{WACV2024RPCANet}. $\gamma$ and $\varepsilon$ are step-size and momentum parameters derived from optimization constants (e.g., Lipschitz constants, $\lambda$, $\mu$). A key limitation is that these parameters are typically \textbf{fixed} across all inputs and iterations.

DRPCA-Net revolutionizes this step with the DTEM. DTEM replaces the fixed parameters and the explicit gradient/proximal operator with learnable components and \textbf{dynamically generated} parameters. Specifically, a lightweight parameter generator network (hypernetwork) predicts stage-specific, input-conditioned parameters $\gamma_D^k$ and $\varepsilon_D^k$. The target update becomes:
\begin{align}
\mathbf{T}^{k} =\ & \gamma_D^{k} \mathbf{T}^{k-1} + (1 - \gamma_D^{k})(\mathbf{D}^{k-1} - \mathbf{B}^{k}) \notag \\
& - \varepsilon_D^{k} \mathcal{S}^{k} \left( \gamma_D^{k} \mathbf{T}^{k-1} + (1 - \gamma_D^{k})(\mathbf{D}^{k-1} - \mathbf{B}^{k}) \right)
\label{eq:dtem_update}
\end{align}
where $S^k(\cdot)$ is a learnable sparsity regularization module instantiated as a small CNN at the $k$-th stage. It serves as a data-driven surrogate for the analytical gradient $\nabla_T S(\cdot)$ in Eq. \eqref{eq:target_update_classical}, learning to approximate its functional effect within the unfolding process. This module corresponds to the DTEM illustrated in Fig.~\ref{fig:drpcanet}, and is responsible for adaptively refining the target feature map by implicitly promoting sparsity based on the input scene.The number of blocks $l_T$ is empirically set to 3 in our experiments. The dynamic parameters $\gamma_D^k$ and $\varepsilon_D^k$, inferred based on scene characteristics (detailed in Sec.~\ref{subsec:dynamic_param_gen}), allow DTEM to adaptively balance information from the previous estimate and the current residual, and to modulate the refinement strength, enhancing robustness across diverse scenarios.

\paragraph{Dynamic Image Reconstruction Module (DIRM)}
While the fundamental RPCA model simply sums the components (Eq. \ref{eq:rpca_decomp}), ensuring fidelity (the third term in Eq.  \ref{eq:rpca_loss}) is crucial. In DRPCA-Net, we introduce the DIRM to potentially refine this process beyond simple addition. DIRM aims to synergistically fuse the estimated background $\mathbf{B}^k$ and target $\mathbf{T}^k$ while potentially enhancing features relevant for subsequent stages or the final output. It is formulated as:
\begin{equation}
\mathbf{D}^{k} = \mathcal{DM}^{k}(\mathbf{B}^{k}, \mathbf{T}^{k}) 
\label{eq:dirm_update}
\end{equation}
where $\mathcal{DM}^k$ represents the reconstruction operation at stage $k$. In our design, $\mathcal{DM}^k$ involves not just summation but also processing by advanced convolutional blocks, such as the proposed Dynamic Residual Group (DRG, detailed in Sec.~\ref{subsec:drg_reconstruction}), to improve the integration and representation quality. This allows for a more sophisticated interaction between the decomposed components than a mere summation.

\subsubsection{Network Architecture Overview}

The overall architecture of DRPCA-Net is depicted in Fig.~\ref{fig:drpcanet}. The network comprises $K$ sequential stages, each structurally identical and implementing one cycle of background estimation, dynamic target extraction, and dynamic image reconstruction. The input infrared image $\mathbf{X}$ initializes $\mathbf{D}^0$, and $\mathbf{T}^0$ is set to zero. The output of stage $k-1$, namely $\mathbf{D}^{k-1}$ and $\mathbf{T}^{k-1}$, are fed into stage $k$. Within stage $k$, the LBEM computes $\mathbf{B}^k$ (Eq. \ref{eq:lbem_update}). Subsequently, the DTEM utilizes $\mathbf{D}^{k-1}$, $\mathbf{B}^k$, and $\mathbf{T}^{k-1}$, along with dynamically generated parameters $(\gamma_D^k, \varepsilon_D^k)$ from the Parameter Generator, to compute $\mathbf{T}^k$ (Eq. \ref{eq:dtem_update}). Finally, the DIRM takes $\mathbf{B}^k$ and $\mathbf{T}^k$ to produce the refined reconstruction $\mathbf{D}^k$ (Eq. \ref{eq:dirm_update}), which serves as input to the next stage. The final target map is typically derived from the target estimate of the last stage, $\mathbf{T}^K$. The entire network, encompassing the parameters of all LBEMs, DTEMs (including $\mathcal{S}^k$), DIRMs (including DRGs), and the Parameter Generator across all $K$ stages, is trained end-to-end.

\subsection{Dynamic Parameter Generation within DTEM}
\label{subsec:dynamic_param_gen}

A salient limitation inherent in conventional deep unfolding architectures is the reliance on static, globally learned scalar parameters (e.g., step sizes, weighting coefficients) that remain invariant across diverse input instances and processing stages. This inherent rigidity significantly impedes the model's capacity to adaptively respond to the complex and heterogeneous characteristics prevalent in infrared imagery, such as varying target saliency, background clutter levels, and noise patterns. To transcend this limitation, DRPCA-Net incorporates a dynamic parameter generation mechanism specifically within the DTEM, enabling stage-wise, input-conditioned modulation of the target refinement process.

Specifically, we instantiate two dedicated, lightweight parameter generator sub-networks, denoted as $P_\gamma$ and $P_\varepsilon$, tasked with inferring the adaptive parameters $\gamma_D^k$ and $\varepsilon_D^k$ (ref. Eq. \eqref{eq:dtem_update}) for each stage $k$. These generators share an identical, computationally efficient architecture designed to distill contextual information into scalar control signals:
\begin{equation}
P(\mathbf{Z}) = \sigma \left( \text{Conv}_2 \left( \delta \left( \text{Conv}_1 \left( \text{GAP}(\mathbf{Z}) \right) \right) \right) \right)
\label{eq:param_gen_arch}
\end{equation}
where $\mathbf{Z}$ represents the input feature map to the generator. $\text{GAP}(\cdot)$ denotes Global Average Pooling, which aggregates spatial information into a global representation. $\text{Conv}_1$ and $\text{Conv}_2$ are $1 \times 1$ convolutional layers, forming a bottleneck structure for efficient channel interaction and dimensionality adjustment. $\delta(\cdot)$ represents the ReLU activation function, and $\sigma(\cdot)$ is the Sigmoid function, ensuring the output parameters reside within the range $(0, 1)$.

Crucially, while sharing the structural blueprint (Eq. \eqref{eq:param_gen_arch}), the two generators are conditioned on distinct, functionally relevant inputs to infer their respective parameters:

\subsubsection{Dynamic Fusion Weight ($\gamma_D^k$)}

This parameter governs the balance between the previous target estimate $\mathbf{T}^{k-1}$ and the current residual information $(\mathbf{D}^{k-1} - \mathbf{B}^{k})$ in the preliminary target update (first line of Eq. \eqref{eq:dtem_update}). Intuitively, $\gamma_D^k$ should reflect the confidence in the previous estimate. Therefore, its generator $P_\gamma$ is conditioned solely on the state of the previous target map:
    \begin{equation}
    \gamma_D^k = P_\gamma\left( \mathbf{T}^{k-1} \right)
    \label{eq:gamma_gen}
    \end{equation}
    This allows the network to dynamically emphasize temporal consistency when the prior target estimate is strong or reliable.

\subsubsection{Dynamic Regularization Strength ($\varepsilon_D^k$)}

This parameter modulates the magnitude of the sparsity-promoting refinement term (second line of Eq. \eqref{eq:dtem_update}), effectively controlling the step size or intensity of the target update based on the sparsity prior $\mathcal{S}^k$. Its value should depend on the characteristics of the \textbf{intermediate} target representation \textbf{before} the sparsity refinement is applied. Consequently, the generator $P_\varepsilon$ is conditioned on this intermediate fused representation:
    \begin{equation}
    \mathbf{T}_{\text{interim}}^k = \gamma_D^k \mathbf{T}^{k-1} + (1 - \gamma_D^k)(\mathbf{D}^{k-1} - \mathbf{B}^{k}) 
    \label{eq:t_interim}
    \end{equation}
    \begin{equation}
    \varepsilon_D^k = P_\varepsilon\left( \mathbf{T}_{\text{interim}}^k \right)
    \label{eq:epsilon_gen}
    \end{equation}
    This design enables the network to adapt the regularization strength based on the current estimate's properties (e.g., intensity, estimated sparsity).

We refer to $P_{\gamma}$ and $P_{\varepsilon}$ as \textit{lightweight hypernetworks}, in the sense that they act as auxiliary, input-conditioned parameter generators that modulate the iterative optimization process. Unlike classical hypernetworks that generate full weight matrices for target networks, the proposed generators produce scalar coefficients that control the update dynamics of each stage. 

The integration of these dynamically generated, input-conditioned parameters $\gamma_D^k$ and $\varepsilon_D^k$ endows the DTEM with significant adaptive capabilities. The dynamic weighting $\gamma_D^k$ facilitates adaptive temporal smoothing, allowing the model to judiciously blend historical information with current observations based on inferred reliability. Concurrently, the dynamic step-size modulation via $\varepsilon_D^k$ enables adaptive regularization, potentially applying stronger refinement when target signals are weak or ambiguous within the residual, and gentler updates when targets are already salient, akin to adaptive learning rate strategies in optimization. This stage-wise, scene-adaptive parameterization is pivotal for enhancing the robustness and generalization performance of DRPCA-Net across diverse and challenging infrared small target detection scenarios.

\subsection{Dynamic Residual Group for Enhanced Reconstruction} 
\label{subsec:drg_reconstruction}

The DIRM, introduced in Eq. \eqref{eq:dirm_update}, serves a critical role beyond the mere summation of the estimated background ($\mathbf{B}^k$) and target ($\mathbf{T}^k$) components. It aims for a synergistic fusion and refinement of these decomposed features to produce an enhanced reconstruction $\mathbf{D}^k$, which benefits subsequent processing stages and the final target extraction. As delineated in Fig.~\ref{fig:drpcanet}, the DIRM employs a bottleneck structure with $1 \times 1$ convolutions for channel dimensionality adjustment, sandwiching its core computational engine: the proposed DRG.

The DRG represents an advancement over conventional residual blocks, such as the Residual Group (RG) structure promulgated in RCAN~\cite{ECCV2018RCAN}. While standard RGs effectively leverage deep residual learning and Channel Attention (CA) mechanisms for feature enhancement, they typically remain agnostic to the spatial variance of feature importance. This limitation can hinder the precise localization and delineation of small, low-contrast targets embedded within complex and spatially heterogeneous infrared backgrounds. To address this, the DRG integrates a novel DSA mechanism alongside CA~\cite{ECCV2018CBAM}, enabling adaptive feature recalibration across both channel and spatial dimensions.

Architecturally, the DRG module processes an input feature map $\mathbf{F}_{\text{in}}$ by passing it through a sequence of $N$ Residual Channel-Spatial Attention Blocks (RCSABs), followed by a final convolutional layer and a long skip connection:
\begin{equation}
\text{DRG}(\mathbf{F}_{\text{in}}) = \mathbf{F}_{\text{in}} + \text{Conv}_{\text{out}} \left( \text{RCSAB}_N(\dots(\text{RCSAB}_1(\mathbf{F}_{\text{in}}))\dots) \right)
\label{eq:drg_structure}
\end{equation}
where $\text{RCSAB}_i$ denotes the $i$-th block, and $\text{Conv}_{\text{out}}$ is typically a $1 \times 1$ convolution for feature integration. This structure facilitates the learning of deep hierarchical features while preserving gradient flow through multiple shortcut connections inherent within the RCSABs and the main DRG body.

\begin{figure}[htbp]
\centering
\includegraphics[scale=0.29]{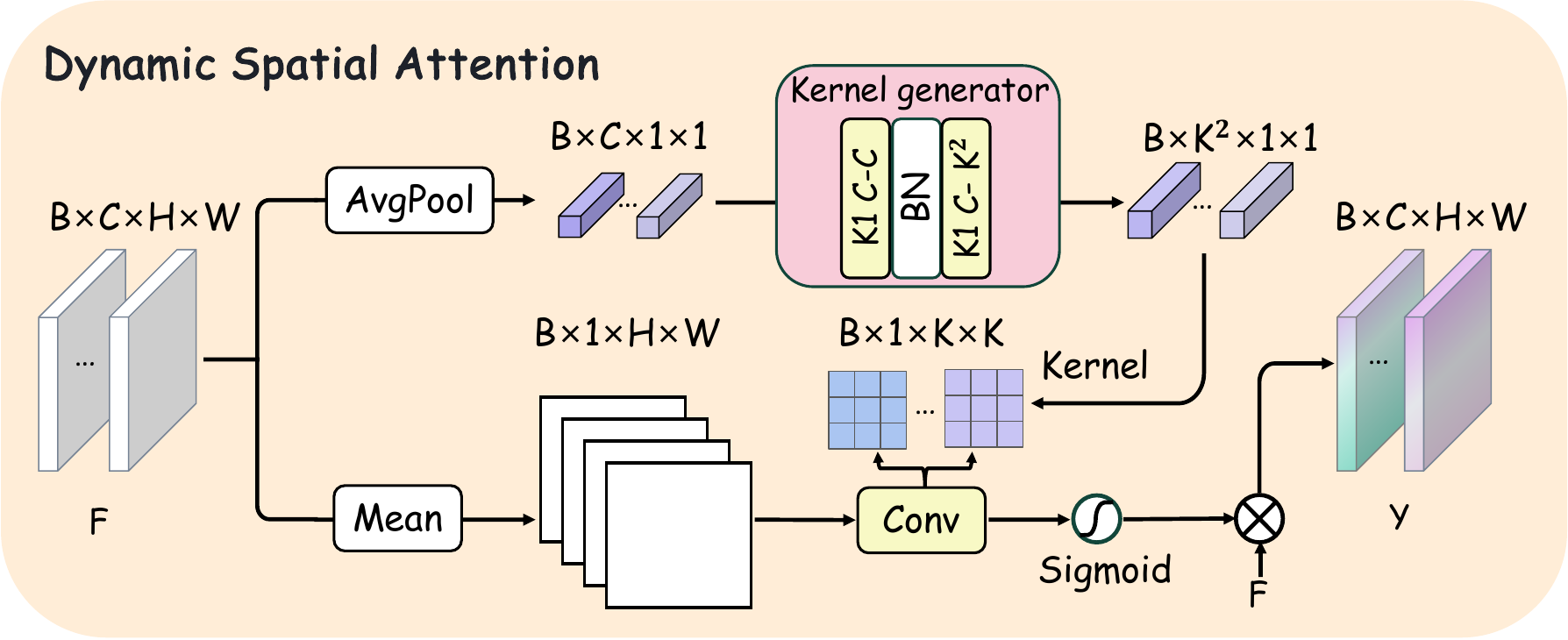}
\caption{Illustration of the proposed Dynamic Spatial Attention (DSA) module. It generates sample-specific convolutional kernels ($W_{ds}$) based on input features ($\mathbf{F}$) to compute an adaptive spatial attention map ($\mathbf{S}_{\text{att}}$), enabling targeted spatial feature modulation for enhanced target preservation during reconstruction.} 
\label{fig:DSA}
\end{figure}

To overcome the spatial insensitivity of channel-centric attention, we introduce the DSA module, illustrated in Fig.~\ref{fig:DSA}. Unlike conventional spatial attention mechanisms that often employ static convolutional kernels or fixed pooling strategies, DSA generates input-conditioned spatial attention maps via dynamically computed convolutional kernels. This allows the network to tailor its spatial focus based on the specific content of each input feature map, thereby achieving bespoke spatial modulation critical for isolating small targets.

Given an intermediate feature map $\mathbf{F} \in \mathbb{R}^{B \times C \times H \times W}$ within an RCSAB, the DSA module operates as follows:
First, a global context vector is derived via Global Average Pooling (GAP) over the spatial dimensions of $\mathbf{F}$. This vector is then processed by a miniature two-layer convolutional network (using $1 \times 1$ convolutions for efficiency) with intermediate ReLU and final Sigmoid activation ($\sigma$) to generate the parameters for the dynamic convolutional kernels:
\begin{equation}
W_{ds} = \sigma \left( \text{Conv}_2 \left( \delta \left( \text{Conv}_1 \left( \text{GAP}(\mathbf{F}) \right) \right) \right) \right)
\label{eq:dsa_kernel_gen}
\end{equation}
where $W_{ds} \in \mathbb{R}^{B \times 1 \times k \times k}$ represents a batch of $k \times k$ convolutional kernels (one per sample in the batch), dynamically generated based on the input $\mathbf{F}$.

Subsequently, a spatial descriptor $\mathbf{F}_{\text{mean}} \in \mathbb{R}^{B \times 1 \times H \times W}$ is obtained by averaging $\mathbf{F}$ across the channel dimension. The dynamic spatial attention map $\mathbf{S}_{\text{att}} \in \mathbb{R}^{B \times 1 \times H \times W}$ is then computed by convolving each sample's spatial descriptor $\mathbf{F}_{\text{mean}}^{(b)}$ with its corresponding dynamically generated kernel $W_{ds}^{(b)}$ (implemented efficiently via grouped convolution):
\begin{equation}
\mathbf{S}_{\text{att}} = \sigma(\text{Conv}_{\text{dynamic}}(\mathbf{F}_{\text{mean}}, W_{ds}))
\label{eq:dsa_map_gen}
\end{equation}
where $\text{Conv}_{\text{dynamic}}$ denotes this sample-wise dynamic convolution operation, and $\sigma$ is the Sigmoid function scaling the attention weights. Finally, the input feature map $\mathbf{F}$ is modulated by these dynamic spatial attention weights:
\begin{equation}
\mathbf{Y} = \mathbf{F} \odot \mathbf{S}_{\text{att}}
\label{eq:dsa_apply}
\end{equation}
where $\odot$ signifies element-wise multiplication (broadcast across the channel dimension). The resulting $\mathbf{Y}$ is the spatially refined feature map.

Although DRPCA-Net introduces learnable modules such as the DRG and dynamic parameter generators, each stage still explicitly follows the decomposition logic of RPCA. The optimization trajectory is preserved in structure, while the learned components enhance adaptability without breaking the interpretability of each sub-step.


\section{Experiments} \label{sec:experiment}

\subsection{Experimental Settings} \label{subsec:setting}

To facilitate a rigorous and comprehensive evaluation of the proposed DRPCA-Net, our experimental protocol leverages established benchmark datasets, standardized evaluation metrics, and meticulously defined implementation parameters.

\subsubsection{Datasets}
We conducted experiments on four widely adopted benchmark datasets for infrared small target detection: SIRST V1~\cite{TGRS2021ACM}, IRSTD-1K~\cite{CVPR2022ISNet}, NUDT-SIRST~\cite{ICCV2019MDvsFAcGAN}, and SIRST-Aug~\cite{TAES2023AGPCNet}. These datasets collectively represent a diverse spectrum of real-world and synthetic infrared imaging scenarios, encompassing significant variations in target scale, intensity, and morphology, as well as background clutter complexity (e.g., urban, maritime, aerial, natural landscapes) and sensor characteristics. The inclusion of both real-world captures and augmented data (SIRST-Aug) ensures a thorough assessment of model effectiveness, robustness, and generalization capabilities. 

\subsubsection{Evaluation Metrics}
In this study, following the mainstream paradigms~\cite{WACV2024RPCANet}. Quantitative performance assessment was carried out using a suite of standard metrics evaluating both pixel-level segmentation accuracy and object-level detection efficacy. Specifically, we report Intersection over Union (IoU) and the F1-measure ($F_1$) to gauge the fidelity of target segmentation masks. For object-level performance, we utilize the Probability of Detection ($P_d$) and False Alarm Rate ($F_a$), computed based on a standard IoU threshold of 0.5 for matching predicted bounding boxes or masks to ground truth targets. Additionally, Receiver Operating Characteristic (ROC) curves are presented to visualize the trade-off between $P_d$ and $F_a$ across different detection thresholds.

\subsubsection{Implementation Details}
To jointly optimize target segmentation and image reconstruction, we adopt a composite loss function defined as:
\begin{equation}
\mathcal{L} = \mathcal{L}_\text{seg} + \lambda \cdot \mathcal{L}_\text{rec}
\label{eq:loss}
\end{equation}
where $\mathcal{L}_\text{seg}$ denotes the SoftIoU loss~\cite{LSP2020SoftIoU}, which directly optimizes the intersection-over-union (IoU) metric and guides the segmentation branch to accurately localize targets. $\mathcal{L}_\text{rec}$ represents the mean squared error (MSE) between the reconstructed image $\mathbf{D}^K$ and the original input image $\mathbf{D}^0$, encouraging the model to preserve background structure and regularize the decomposition process. The weighting coefficient $\lambda$ balances the contributions of the two terms. Following empirical evaluation in Sec.~\ref{subsubsec:loss_lambda_ablation}, we set $\lambda = 0.1$, which offers a favorable trade-off between performance and stability.

Optimization was performed using the Adam optimizer with an initial learning rate set to $1 \times 10^{-4}$. The learning rate decayed throughout training following a polynomial schedule defined by $(1 - \text{iter}/\text{total\_iter})^{0.9}$. A batch size of 8 was used across all experiments. Models were trained for a total of 400 epochs. All experiments were implemented within the PyTorch framework and executed on NVIDIA GeForce RTX 4090 GPUs. For establishing comparative benchmarks, we utilized the publicly available BasicISTD Toolbox v2.0 where applicable; for methods not included therein, we meticulously followed their officially released codebases and training protocols to ensure equitable comparison.

\subsection{Ablation Study} \label{subsec:ablation}

To rigorously validate the architectural design of DRPCA-Net and quantify the contributions of its constituent components, we performed a series of systematic ablation experiments. These studies dissect the impact of the dynamic parameter generation mechanism, the DRG, and key hyperparameters. Evaluations were conducted on the SIRST V1 and NUDT-SIRST benchmark datasets, with results presented in Tab. \ref{tab:ablation}-\ref{tab:lambda_ablation}. Optimal performance values are marked with red font, and \( \uparrow \) / \( \downarrow \) indicate preference for higher/lower metric values, respectively.

\subsubsection{Efficacy of Core Components}
We first assessed the individual and synergistic contributions of the primary innovations within DRPCA-Net: the dynamic parameter generators ($P_\gamma, P_\varepsilon$) and the DRG. Starting from a baseline unfolding network employing static learnable parameters and standard convolutional reconstruction (Tab.~\ref{tab:ablation}, Strategy (a)), we incrementally incorporated these modules. Introducing the dynamic $\varepsilon$-generator (Strategy (b)) and subsequently the $\gamma$-generator (Strategy (c)) yielded progressive performance improvements, confirming the benefits of adaptive, input-conditioned parameterization over static counterparts. Independently replacing the standard reconstruction block with our DRG module (Strategy (d)) also resulted in substantial gains, highlighting its superior feature fusion and refinement capabilities. The full DRPCA-Net configuration (Strategy (f)), integrating both dynamic parameter generation and the DRG, achieved the highest performance, demonstrating a significant synergistic effect between adaptive optimization dynamics and enhanced feature representation (e.g., +7.15\% mIoU on SIRST V1 vs. baseline). This validates the effectiveness of our integrated design philosophy.

\begin{table}[h]
    \centering
    \caption{Quantitative ablation study assessing the impact of core components, including \(\varepsilon\)-generator, \(\gamma\)-generator, and DRG.  The best results are highlighted in bold red font.}
    \label{tab:ablation}
    \renewcommand{\arraystretch}{1.3} 
    \setlength{\tabcolsep}{4pt} 
    \footnotesize 
    \begin{tabular}{c|ccc|cc|cc}
        \toprule
        \multirow{2}{*}{\textbf{Strategy}} & \multicolumn{3}{c|}{\textbf{Module}} & \multicolumn{2}{c|}{\textbf{SIRST V1}} & \multicolumn{2}{c}{\textbf{NUDT-SIRST}} \\
        & $\varepsilon$-Gen & $\gamma$-Gen & DRG & mIoU$\uparrow$ & F$_1\uparrow$ & mIoU$\uparrow$ & F$_1\uparrow$ \\
        \midrule
        (a) & \ding{55} & \ding{55} & \ding{55} & 68.37 & 81.21 & 88.24 & 93.75 \\ 
        \rowcolor{gray!10} 
        (b) & \ding{51} & \ding{55} & \ding{55} & 70.36 & 82.60 & 90.25 & 94.87 \\  
        (c) & \ding{55} & \ding{51} & \ding{55} & 70.43 & 82.65 & 89.93 & 94.69 \\  
        \rowcolor{gray!10} 
        (d) & \ding{55} & \ding{55} & \ding{51} & 73.69 & 84.85 & 93.47 & 96.62 \\  
        (e) & \ding{51} & \ding{51} & \ding{55} & 74.74 & 85.55 & 91.27 & 94.86 \\ 
        \rowcolor{RoyalBlue!8} 
        (f) & \ding{51} & \ding{51} & \ding{51} & \textbf{\textcolor[rgb]{1,0,0}{75.52}} & \textbf{\textcolor[rgb]{1,0,0}{86.05}} & \textbf{\textcolor[rgb]{1,0,0}{94.16}} & \textbf{\textcolor[rgb]{1,0,0}{96.99}} \\  
        \bottomrule
    \end{tabular}
\end{table}

\begin{table}[h]
    \centering
    \caption{Ablation study on the parameter generator module across different configurations.  The best results are highlighted in bold red font.}
    \label{tab:param_gen}
    \renewcommand{\arraystretch}{1.3}
    \setlength{\tabcolsep}{5pt} 
    \begin{tabular}{l|cc|cc}
        \toprule
        \multicolumn{1}{c|}{\multirow{2}{*}{\textbf{Strategy}}} & \multicolumn{2}{c|}{\textbf{SIRST V1}} & \multicolumn{2}{c}{\textbf{NUDT-SIRST}} \\
        & mIoU \( \uparrow \) & F$_1\uparrow$ & mIoU \( \uparrow \) & F$_1\uparrow$ \\
        \midrule
        (a) Learnable  & 67.47 & 80.58 & 86.49 & 92.76 \\  
        \rowcolor{gray!10}
        (b) CBAM       & 74.31 & 85.26 & 57.87 & 73.32 \\  
        (c) Simple     & 72.72 & 84.20 & 89.77 & 94.60 \\  
        \rowcolor{gray!10}
        (d) FC         & 72.39 & 83.99 & 91.06 & 95.23 \\
        (e) LSK        & 68.67 & 81.43 & 87.65 & 93.43 \\
        \rowcolor{gray!10}
        (f) CBAM(map)  & 67.87 & 80.86 & 82.99 & 90.70 \\
        (g) Multiscale  & 65.82 & 79.38 & 86.81 & 92.94 \\
        \rowcolor{RoyalBlue!8}
        (h) \textbf{Ours}       & \textbf{\textcolor[rgb]{1,0,0}{74.74}} & \textbf{\textcolor[rgb]{1,0,0}{85.55}} & \textbf{\textcolor[rgb]{1,0,0}{91.27}} & \textbf{\textcolor[rgb]{1,0,0}{95.43}} \\  
        \bottomrule
    \end{tabular}
\end{table}

\subsubsection{Analysis of Dynamic Parameter Generation}
The design of the parameter generator itself is critical. We compared our proposed lightweight architecture (Tab.~\ref{tab:param_gen}, Strategy (h)) against several alternatives: static learnable scalars (a), simpler scalar regression mappings (single Conv layer (c), FC layer (d)), attention-based structures (CBAM (b)), and three designs that attempt to generate spatial parameter maps rather than scalar coefficients — namely LSK (e), CBAM spatial attention (f), and a newly introduced multi-scale structure (g).
Strategies (e), (f), and (g) all aim to produce spatially varying parameter maps as opposed to global scalar coefficients. Strategy (e) uses the LSK attention module to generate location-sensitive modulation maps; Strategy (f) directly adopts the spatial attention map from CBAM; and Strategy (g) employs a multi-branch convolutional structure with kernel sizes 1, 3, and 5, followed by channel fusion to compute a pixel-wise attention map. However, despite their higher representational capacity, these approaches exhibited inferior performance compared to scalar-based generators. Spatial modulation introduces unnecessary complexity and noise when applied to global optimization parameters like step sizes and momentum coefficients, which inherently require global semantic understanding rather than local spatial detail.
By contrast, scalar-generating designs — including strategies (b), (c), and (d) — are more aligned with the theoretical formulation of RPCA, where the update coefficients are global constants. Among them, our proposed architecture (h) achieves consistently superior performance across datasets, benefiting from its ability to extract compact, globally representative features via GAP and lightweight convolutions. Static parameters (a) proved insufficient, lacking adaptability to varying scene statistics, while simpler mappings (c, d) showed limited representational power.
Furthermore, an ablation study on the number of intermediate channels (Tab.~\ref{tab:channel_ablation}) confirmed that using 3 channels offers an ideal trade-off between model capacity and generalization.

\begin{table}[h]
    \centering
    \caption{Ablation Study on the Number of Channels in the Parameter Generator.  The best results are highlighted in bold red font.}
    \label{tab:channel_ablation}
    \renewcommand{\arraystretch}{1.3} 
    \setlength{\tabcolsep}{4pt}
    \begin{tabular}{c|cc|cc}
        \toprule
        \multirow{2}{*}{\textbf{Channels}} & \multicolumn{2}{c|}{\textbf{SIRST V1}} & \multicolumn{2}{c}{\textbf{NUDT-SIRST}} \\
        & mIoU \( \uparrow \) & F$_1\uparrow$ & mIoU \( \uparrow \) & F$_1\uparrow$ \\
        \midrule
        \rowcolor{RoyalBlue!5}
        2 & 70.83 & 82.92 & 90.26 & 94.88 \\  
        \rowcolor{RoyalBlue!8}
        \textbf{3} & \textbf{\textcolor[rgb]{1,0,0}{74.74}} & \textbf{\textcolor[rgb]{1,0,0}{85.69}} & \textbf{\textcolor[rgb]{1,0,0}{91.57}} & \textbf{\textcolor[rgb]{1,0,0}{95.60}} \\
        \rowcolor{RoyalBlue!11}
        4 & 72.19 & 83.58 & 91.00 & 95.28 \\ 
        \rowcolor{RoyalBlue!14}
        8 & 73.87 & 84.97 & 90.20 & 94.85 \\  
        \bottomrule
    \end{tabular}
\end{table}

\subsubsection{Analysis of the Dynamic Residual Group (DRG)}
We investigated the internal configuration of the DRG by evaluating different attention mechanisms within its RCSABs (Tab.~\ref{tab:drg_ablation}). Comparing enhanced channel attention (CA2, based on CBAM) against a simpler baseline (CA1), CA2 demonstrated superior inter-channel relationship modeling (cf. Strategy (a) vs. (b)). Crucially, comparing our proposed DSA against standard Spatial Attention (SA) revealed the significant advantage of dynamic, input-conditioned spatial modulation for small target detection (cf. Strategy (c) vs. (d), showing +0.67\% mIoU for DSA on SIRST V1). The optimal configuration (Strategy (d)) integrates both CA2 and DSA, confirming the effectiveness of coordinating adaptive attention across both channel and spatial dimensions for refined target feature extraction and reconstruction. Additionally, varying the number of RCSABs within the DRG (Tab.~\ref{tab:rcsab_ablation}) indicated that $N=5$ blocks provide the optimal depth, balancing sufficient hierarchical feature refinement against computational cost and the risk of diminishing returns or overfitting associated with excessive depth.

\begin{table}[h]
    \centering
    \caption{Ablation study on the DRG structure. The best results are highlighted in bold red font.}
    \label{tab:drg_ablation}
    \renewcommand{\arraystretch}{1.3} 
    \setlength{\tabcolsep}{4pt} 
    \footnotesize 
    \begin{tabular}{c|cccc|cc|cc}
        \toprule
        \multirow{2}{*}{\textbf{DRG}} & \multicolumn{4}{c|}{\textbf{Attention Modules}} & \multicolumn{2}{c|}{\textbf{SIRST V1}} & \multicolumn{2}{c}{\textbf{NUDT-SIRST}} \\
        & CA1 & CA2 & SA & DSA & mIoU$\uparrow$ & F$_1\uparrow$ & mIoU$\uparrow$ & F$_1\uparrow$ \\
        \midrule
        (a) & \ding{51} & \ding{55} & \ding{55} & \ding{55} & 71.62 & 83.46 & 91.76 & 95.70 \\  
        \rowcolor{gray!10}
        (b) & \ding{55} & \ding{51} & \ding{55} & \ding{55} & 72.01 & 83.72 & 92.19 & 95.93 \\  
        (c) & \ding{55} & \ding{51} & \ding{51} & \ding{55} & 72.58 & 84.11 & 92.80 & 96.26 \\  
        \rowcolor{RoyalBlue!8}
        (d) & \ding{55} & \ding{51} & \ding{55} & \ding{51} & \textbf{\textcolor[rgb]{1,0,0}{73.69}} & \textbf{\textcolor[rgb]{1,0,0}{84.85}} & \textbf{\textcolor[rgb]{1,0,0}{93.47}} & \textbf{\textcolor[rgb]{1,0,0}{96.62}} \\  
        \bottomrule
    \end{tabular}
\end{table} 

\begin{table}[h]
    \centering
    \caption{Ablation study on the number of RCSABs. The best results are highlighted in bold red font.}
    \label{tab:rcsab_ablation}
    \renewcommand{\arraystretch}{1.3}
    \setlength{\tabcolsep}{4.5pt}
    \begin{tabular}{c|c|cc|cc}
        \toprule
        \textbf{RCSAB} & \multirow{2}{*}{\textbf{Params}} & \multicolumn{2}{c|}{\textbf{SIRST V1}} & \multicolumn{2}{c}{\textbf{NUDT-SIRST}} \\
        \textbf{Number} &                 & mIoU $\uparrow$ & F$_1\uparrow$ & mIoU $\uparrow$ & F$_1\uparrow$ \\
        \midrule
        \rowcolor{RoyalBlue!2}
        3 & 0.927M & 72.23 & 83.87 & 91.66 & 95.65 \\ 
        \rowcolor{RoyalBlue!5}
        4 & 1.048M & 73.03 & 84.43 & 92.37 & 96.03 \\  
        \rowcolor{RoyalBlue!8}
        \textbf{5} & 1.169M & \textbf{\textcolor[rgb]{1,0,0}{73.69}} & \textbf{\textcolor[rgb]{1,0,0}{84.85}} & \textbf{\textcolor[rgb]{1,0,0}{93.47}} & \textbf{\textcolor[rgb]{1,0,0}{96.62}} \\  
        \rowcolor{RoyalBlue!11}
        6 & 1.290M & 72.10 & 83.79 & 92.92 & 96.33 \\  
        \bottomrule
    \end{tabular}
\end{table}

\subsubsection{Impact of Unfolding Depth}
Finally, we examined the influence of the number of unfolding stages ($K$) on DRPCA-Net's performance (Tab.~\ref{tab:stage_ablation}). Increasing the number of stages from 4 to 6 progressively improved performance, suggesting that deeper unfolding allows for more thorough iterative refinement of the background and target components, better approximating the RPCA optimization process. However, increasing the depth further to 7 stages led to a slight performance degradation. This indicates that $K=6$ stages represent an optimal trade-off, providing sufficient modeling capacity without incurring excessive computational overhead or potential optimization instability associated with overly deep unfolded networks.

\begin{table}[h]
    \centering
    \caption{Ablation study on the number of unfolding stages in DRPCA-Net. The best results are highlighted in bold red font.}
    \label{tab:stage_ablation}
    \renewcommand{\arraystretch}{1.3}
    \setlength{\tabcolsep}{4.5pt}
    \begin{tabular}{c|c|cc|cc}
        \toprule
        \textbf{Stages} & \multirow{2}{*}{\textbf{Params}} & \multicolumn{2}{c|}{\textbf{SIRST V1}} & \multicolumn{2}{c}{\textbf{NUDT-SIRST}} \\
        \textbf{Number} &                 & mIoU $\uparrow$ & F$_1\uparrow$ & mIoU $\uparrow$ & F$_1\uparrow$ \\
        \midrule
        \rowcolor{RoyalBlue!2}
        4 & 0.779M & 72.61 & 84.13 & 93.09 & 96.42 \\
        \rowcolor{RoyalBlue!5}
        5 & 0.974M & 74.32 & 85.27 & 93.48 & 96.63 \\
        \rowcolor{RoyalBlue!8}
        \textbf{6} & 1.169M & \textbf{\textcolor[rgb]{1,0,0}{75.52}} & \textbf{\textcolor[rgb]{1,0,0}{86.05}} & \textbf{\textcolor[rgb]{1,0,0}{94.16}} & \textbf{\textcolor[rgb]{1,0,0}{96.99}} \\
        \rowcolor{RoyalBlue!11}
        7 & 1.364M & 73.89 & 84.99 & 93.57 & 96.68 \\
        \bottomrule
    \end{tabular}
\end{table}

\subsubsection{Impact of Reconstruction Loss Weight $\lambda$}
\label{subsubsec:loss_lambda_ablation}
We evaluate the impact of the reconstruction loss weight $\lambda$ in the total loss function (Eq. \eqref{eq:loss}). As shown in Tab.\ref{tab:lambda_ablation}, setting $\lambda = 0.1$ achieves the best performance on both SIRST V1 and NUDT-SIRST datasets. Larger values (e.g., 0.5) overemphasize reconstruction, while smaller values (e.g., 0.05) reduce regularization. Based on this, we adopt $\lambda = 0.1$ as the default setting.

\begin{table}[h]
    \centering
    \caption{Ablation study on the reconstruction loss weight $\lambda$ in the total loss function.The best results are highlighted in bold red font.}
    \label{tab:lambda_ablation}
    \renewcommand{\arraystretch}{1.3} 
    \setlength{\tabcolsep}{4pt}
    \begin{tabular}{c|cc|cc}
        \toprule
        \multirow{2}{*}{\textbf{Weight $\lambda$}} & \multicolumn{2}{c|}{\textbf{SIRST V1}} & \multicolumn{2}{c}{\textbf{NUDT-SIRST}} \\
        & mIoU \( \uparrow \) & F$_1\uparrow$ & mIoU \( \uparrow \) & F$_1\uparrow$ \\
        \midrule
        \rowcolor{RoyalBlue!5}
        0.5 & 73.87 & 84.91 & 92.80 & 96.27 \\  
        \rowcolor{RoyalBlue!8}
        \textbf{0.1} & \textbf{\textcolor[rgb]{1,0,0}{75.52}} & \textbf{\textcolor[rgb]{1,0,0}{86.05}} & \textbf{\textcolor[rgb]{1,0,0}{94.16}} & \textbf{\textcolor[rgb]{1,0,0}{96.99}} \\
        \rowcolor{RoyalBlue!11}
        0.05 & 72.54 & 72.97 & 93.83 & 96.82 \\ 
        \bottomrule
    \end{tabular}
\end{table}

\renewcommand{\arraystretch}{1.2} 
\begin{table*}[htbp]
    \centering
    \footnotesize 
    \setlength{\tabcolsep}{3.5pt} 
    \caption{Performance comparison of different methods on SIRST V1, NUDT-SIRST, SIRST-Aug, and IRSTD-1K in terms of IoU (\%), F$_1$ (\%), P$_d$ (\%), F$_a$ (10$^{-6}$), and the number of parameters (M). Evaluations are conducted on 256 × 256 images. The best results are highlighted in bold red font.}
    \label{tab:comparison}
    \begin{tabular}{l|c|cccc|cccc|cccc|cccc}
        \toprule
        \multicolumn{1}{c|}{\multirow{2}{*}{\textbf{Methods}}} & \multirow{2}{*}{\textbf{Params}} & \multicolumn{4}{c|}{\textbf{SIRST V1}} & \multicolumn{4}{c|}{\textbf{NUDT-SIRST}} & \multicolumn{4}{c|}{\textbf{SIRST-Aug}} & \multicolumn{4}{c}{\textbf{IRSTD-1K (Failure Case)}} \\
        & & IoU$\uparrow$ & F$_1\uparrow$ & P$_d\uparrow$ & F$_a\downarrow$ & IoU$\uparrow$ & F$_1\uparrow$ & P$_d\uparrow$ & F$_a\downarrow$ & IoU$\uparrow$ & F$_1\uparrow$ & P$_d\uparrow$ & F$_a\downarrow$ & IoU$\uparrow$ & F$_1\uparrow$ & P$_d\uparrow$ & F$_a\downarrow$ \\
        \midrule
        \multicolumn{5}{@{}l}{\textit{Traditional Method}} \\
        \midrule
        Tophat~\cite{RP2010TopHat} & - & 26.93 & 42.44 & 88.99 & 182.7 & 22.25 & 36.41 & 91.21 & 757.3 & 16.37 & 28.14 & 81.15 & 160.3 & 6.64 & 12.46 & 81.09 & 142.4 \\
        \rowcolor{gray!10} 
        Max-Median~\cite{SPIE1999DeshpandeMaxMean} & - & 32.56 & 49.12 & 94.46 & 33.88 & 20.49 & 34.01 & 84.97 & 312.7 & 16.88 & 28.89 & 79.36 & 82.57 & 20.76 & 34.38 & 85.56 & 212.1 \\
        MPCM~\cite{RP2016Multiscale} & - & 22.40 & 36.60 & 88.07 & 408.3 & 9.25 & 16.94 & 67.33 & 317.1 & 19.76 & 33.00 & 90.92 & 31.42 & 15.52 & 26.88 & 61.85 & 109.3 \\
        \rowcolor{gray!10} 
        HBMLCM~\cite{LGRS2018HBMLCM} & - & 11.95 & 21.36 & 79.81 & 413.2 & 7.60 & 14.13 & 60.74 & 178.4 & 2.37 & 46.37 & 50.61 & 1461 & 12.60 & 22.38 & 57.73 & 100.8 \\
        NWMTH~\cite{RP2010TopHat} & - & 27.22 & 42.80 & 88.99 & 25.37 & 14.36 & 25.11 & 71.42 & 199.1 & 13.31 & 23.50 & 83.76 & 45.72 & 14.54 & 25.39 & 76.97 & 24.75 \\
        \rowcolor{gray!10} 
        FKRW~\cite{TGRS2019FKRW} & - & 10.43 & 18.89 & 78.89 & 15.26 & 9.37 & 17.13 & 65.29 & 86.27 & 4.12 & 7.93 & 60.11 & 119.9 & 4.94 & 9.41 & 58.41 & 21.10 \\
        TLLCM~\cite{LGRS2019TLLCM} & - & 18.73 & 31.55 & 81.65 & 21.11 & 12.33 & 21.96 & 72.63 & 119.3 & 2.60 & 5.08 & 24.89 & 18.28 & 10.24 & 18.58 & 64.60 & 10.40 \\
        \rowcolor{gray!10} 
        IPI~\cite{TIP2013IPI} & - & 32.30 & 48.83 & 95.41 & 392.6 & 31.71 & 48.16 & 87.30 & 109.4 & 19.00 & 31.93 & 75.79 & 101.9 & 26.55 & 41.96 & 89.34 & 71.63 \\
        PSTNN~\cite{RS2019PSTNN} & - & 36.28 & 53.24 & 86.23 & 74.16 & 21.69 & 35.64 & 68.04 & 222.1 & 17.75 & 30.15 & 58.59 & 92.05 & 21.35 & 35.19 & 68.04 & 270.1 \\
        \rowcolor{gray!10} 
        NRAM~\cite{RS2019NRAM} & - & 17.16 & 29.29 & 71.13 & 33.48 & 12.70 & 22.54 & 63.49 & 27.37 & 8.56 & 15.77 & 58.59 & 46.81 & 21.05 & 34.79 & 78.89 & 10.47 \\
        NOLC~\cite{RS2019NOLC} & - & 20.35 & 33.82 & 83.48 & 13.66 & 18.60 & 31.37 & 70.68 & 39.69 & 8.68 & 15.97 & 59.00 & 66.49 & 10.90 & 19.65 & 64.26 & 18.30 \\
        \rowcolor{gray!10} 
        NIPPS~\cite{Jinfrared2017NIPPS} & - & 28.36 & 44.18 & 81.65 & 20.94 & 19.11 & 32.99 & 69.52 & 41.46 & 9.78 & 17.82 & 52.54 & 83.66 & 21.08 & 34.82 & 76.88 & 22.24 \\
        \midrule
        \multicolumn{5}{@{}l}{\textit{Deep Learning Methods}} \\
        \midrule
        ACM~\cite{TGRS2021ACM} & 0.398 & 69.67 & 82.12 & 98.16 & 26.79 & 69.22 & 81.79 & 94.39 & 12.85 & 71.92 & 83.67 & 94.63 & 204.4 & 54.35 & 70.26 & 84.53 & 11.01 \\
        \rowcolor{gray!10} 
        ALCNet~\cite{TGRS2021ALCNet} & 0.378 & 67.73 & 80.85 & 99.08 & 7.10 & 69.14 & 81.76 & 96.50 & 30.82 & 72.94 & 84.35 & 98.07 & 54.20 & 57.28 & 72.71 & 85.56 & 13.36 \\
        ISNet~\cite{CVPR2022ISNet} & 0.967 & 67.11 & 80.31 & 99.08 & 57.66 & 87.53 & 93.28 & 96.50 & 12.20 & 72.04 & 83.74 & 96.14 & \textbf{\textcolor[rgb]{1,0,0}{27.13}} & 63.42 & 76.77 & 88.62 & 28.24 \\
        \rowcolor{gray!10} 
        RDIAN~\cite{TGRS2023IRDST} & 0.216 & 68.76 & 81.52 & 98.16 & 36.02 & 85.24 & 92.03 & 97.03 & 14.27 & 74.88 & 85.63 & 98.07 & 53.59 & 63.59 & 77.64 & 88.31 & 12.98 \\
        ResUNet~\cite{ISPRS2020ResUNet}& 0.904 & 72.31 & 83.94 & 97.24 & 7.63 & 90.97 & 95.23 & 96.08 & 5.93 & 72.23 & 83.87 & 97.79 & 41.32 & 63.08 & 77.33 & 92.09 & 19.05 \\
        \rowcolor{gray!10} 
        ISTDU-Net~\cite{LGRS2022ISTDUNet} & 2.751 & 73.73 & 84.88 & 99.08 & 8.87 & 92.99 & 96.36 & 98.09 & 3.98 & 74.79 & 85.56 & 97.11 & 30.69 & 65.25 & 79.04 & 91.40 & 19.74 \\
        DNANet~\cite{TIP2023DNANet} & 4.697 & 73.03 & 84.44 & 99.08 & 2.46 & 93.00 & 96.37 & 97.98 & 9.10 & 71.96 & 83.68 & 96.28 & 42.33 & 65.52 & 79.24 & \textbf{\textcolor[rgb]{1,0,0}{93.47}} & 7.82 \\
        \rowcolor{gray!10} 
        MTUNet~\cite{TGRS2023MTUNet}& 8.220 & 73.71 & 84.89 & 98.16 & 1.06 & 91.60 & 95.58 & 97.56 & 6.48 & 73.18 & 84.50 & 98.34 & 58.71 & 63.65 & 77.74 & 91.75 & 14.88 \\
        AGPCNet~\cite{TAES2023AGPCNet}& 12.36 & 68.84 & 81.54 & 98.16 & 17.57 & 82.52 & 90.43 & 96.71 & 13.42 & 74.07 & 85.11 & 98.07 & 32.67 & 60.18 & 75.14 & 90.72 & 13.44 \\
        \rowcolor{gray!10} 
        UIUNet~\cite{TIP2023UIUNet}& 50.54 & 74.91 & 85.63 & 98.16 & \textbf{\textcolor[rgb]{1,0,0}{0.53}} & 92.51 & 96.08 & 96.50 & \textbf{\textcolor[rgb]{1,0,0}{1.70}} & 73.45 & 84.69 & 97.79 & 40.90 & 65.61 & 79.17 & 90.03 & 6.98 \\
        MSHNet~\cite{CVPR2024MSHNet}& 4.066 & 68.35 & 81.20 & 98.16 & 28.92 & 77.92 & 87.59 & 96.82 & 16.70 & 72.97 & 84.33 & 93.25 & 305.3 & 64.56 & 78.56 & 89.45 & \textbf{\textcolor[rgb]{1,0,0}{7.21}}  \\
        \rowcolor{gray!10}
        RPCANet~\cite{WACV2024RPCANet} & 0.680 & 68.37 & 81.21 & 94.49 & 24.49 & 88.24 & 93.75 & 96.40 & 20.96 & 71.64 & 83.48 & 97.93 & 97.88 & 61.23 & 75.95 & 89.69 & 22.48 \\
        L$^{2}$SKNet~\cite{TGRS2025L2SKNet} & 0.899 & 72.23 & 83.87 & 98.16 & 9.23 & 93.13 & 96.44 & 98.09 & 5.81 & 72.75 & 84.22 & 96.56 & 54.82 & 63.73 & 77.84 & 92.78 & 16.85 \\
        \rowcolor{gray!10} 
        SCTransNet~\cite{TGRS2025SCTransNet} & 11.19 & 71.15 & 83.14 & 97.24 & 8.87 & 89.49 & 94.45 & 97.67 & 10.29 & 72.75 & 84.20 & 97.76 & 52.04 & 62.04 & 76.56 & 73.47 & 23.68 \\
        PConv~\cite{AAAI2025pinwheel} & 4.064 & 72.62 & 84.14 & 99.08 & 21.46 & 78.58 & 88.00 & 97.56 & 14.13 & 73.55 & 84.76 & 97.93 & 74.81 & \textbf{\textcolor[rgb]{1,0,0}{67.45}} & \textbf{\textcolor[rgb]{1,0,0}{80.56}} & 93.19 & 9.71 \\
        \rowcolor{RoyalBlue!8}
        \textbf{Ours} & 1.169 & \textbf{\textcolor[rgb]{1,0,0}{75.52}} & \textbf{\textcolor[rgb]{1,0,0}{86.05}} & \textbf{\textcolor[rgb]{1,0,0}{99.08}} & 3.73 & \textbf{\textcolor[rgb]{1,0,0}{94.16}} & \textbf{\textcolor[rgb]{1,0,0}{96.99}} & \textbf{\textcolor[rgb]{1,0,0}{98.41}} & 2.55 & \textbf{\textcolor[rgb]{1,0,0}{76.50}} & \textbf{\textcolor[rgb]{1,0,0}{86.69}} & \textbf{\textcolor[rgb]{1,0,0}{98.62}} & 30.60 & 64.14 & 78.15 & 92.09 & 17.92 \\
        \bottomrule
    \end{tabular}
\end{table*}

\subsection{Comparison with State-of-the-Arts} \label{subsec:sota}

To ascertain the efficacy and advancement offered by DRPCA-Net, we conducted a rigorous comparative analysis against a wide array of state-of-the-art (SOTA) infrared small target detection algorithms. This benchmark suite encompasses both classical model-driven techniques (e.g., Top-hat~\cite{RP2010TopHat}, IPI~\cite{TIP2013IPI}, PSTNN~\cite{RS2019PSTNN}) and contemporary data-driven deep learning frameworks, including recent end-to-end CNNs and other deep unfolding networks (e.g., ACM~\cite{TGRS2021ACM}, ISNet~\cite{CVPR2022ISNet}, UIUNet~\cite{TIP2023UIUNet}, L$^2$SKNet~\cite{TGRS2025L2SKNet}, RPCANet~\cite{WACV2024RPCANet},PConv~\cite{AAAI2025pinwheel}). Notably, the PConv model is an enhanced version of MSHNet, in which a pinwheel-shaped convolution is introduced to improve feature representation. Performance was evaluated across the four aforementioned public datasets using standard metrics: mIoU, $F_1$, $P_d$, and $F_a$. Detailed quantitative results are presented in Tab.~\ref{tab:comparison}.

The empirical evidence underscores the superior performance of DRPCA-Net. As delineated in Tab.~\ref{tab:comparison}, our proposed method consistently achieves state-of-the-art or highly competitive results across the majority of datasets and evaluation metrics. Notably, DRPCA-Net secures the top rank in mIoU, $F_1$, and $P_d$ on the SIRST V1, SIRST-AUG, and NUDT-SIRST datasets. This consistent superiority attests to the potency of the dynamic unfolding mechanism and the enhanced feature representation facilitated by the DRG module, enabling robust adaptation to diverse target characteristics and background complexities. While traditional model-driven approaches exhibit inherent limitations in handling complex scenes, often yielding lower segmentation accuracy and higher false alarms, deep learning paradigms demonstrate significantly improved efficacy. Within this advanced category, DRPCA-Net further distinguishes itself.

Furthermore, DRPCA-Net achieves this state-of-the-art performance with remarkable computational efficiency. Possessing only 1.169 M parameters, it strikes an advantageous balance between detection prowess and model complexity, rendering it practical for resource-constrained applications.

\begin{figure}[htbp]
\centering
\includegraphics[scale=0.44]{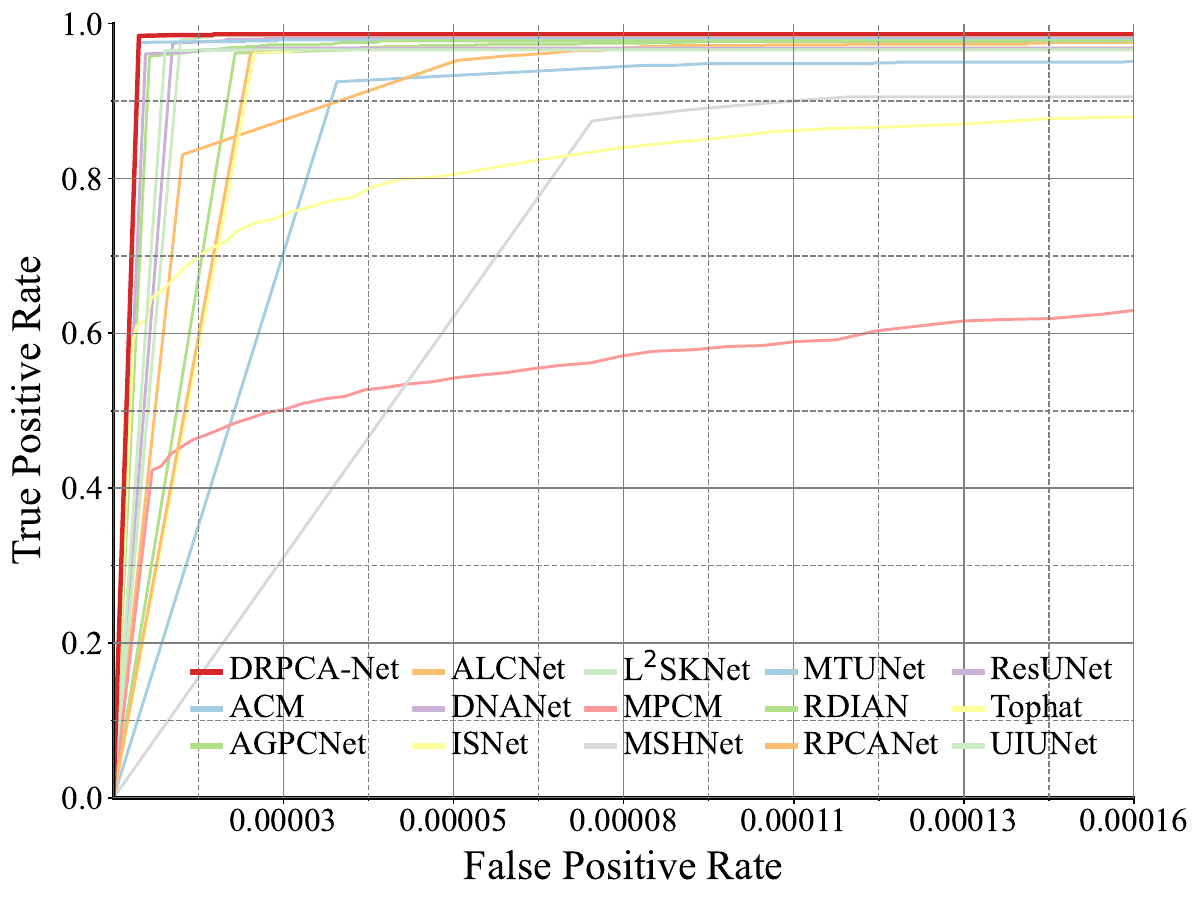}
\caption{Comparative ROC curve analysis of detection methods on NUDT-SIRST dataset.}
\label{fig:ROCNUDT}
\end{figure}

The quantitative superiority of DRPCA-Net is visually corroborated by the ROC analysis presented in Fig.~\ref{fig:ROCNUDT} for the NUDT-SIRST dataset. The ROC curve for DRPCA-Net exhibits a significantly steeper ascent towards the top-left corner compared to nearly all other evaluated methods. This trajectory signifies a higher $P_d$ achievable for any given $F_a$, particularly under stringent low $F_a$ regimes. Such characteristics underscore the model's enhanced robustness against background clutter and its heightened sensitivity in discerning faint or challenging targets, which are critical capabilities in practical infrared detection scenarios.

In conclusion, the comprehensive comparative evaluation confirms that DRPCA-Net advances the state of the art in infrared small target detection, offering superior accuracy, robustness, and a favorable efficiency profile, largely attributable to its principled integration of dynamic adaptation within the deep unfolding RPCA framework.

\subsection{\textbf{Discussion on Failure Cases and Model Priors}} 

Notwithstanding its overall state-of-the-art performance, a nuanced observation emerges from Tab.~\ref{tab:comparison}: DRPCA-Net demonstrates particularly dominant results on the SIRSTV1, NUDT-SIRST, and SIRST-AUG datasets, while exhibiting comparatively less, though still highly competitive, performance ascendancy on IRSTD-1K. We posit that this variation stems fundamentally from the degree of congruence between the intrinsic characteristics of each dataset and the canonical low-rank background plus sparse target assumption underpinning the Robust PCA model, which DRPCA-Net dynamically unfolds. The former three datasets predominantly feature aero-spatial scenes, where backgrounds often exhibit strong spatial correlations amenable to low-rank approximation, and targets manifest as sparse, localized anomalies, thereby aligning well with the RPCA prior. Conversely, the IRSTD-1K dataset incorporates a significant proportion of complex terrestrial scenes replete with salient clutter and interference that can visually mimic small targets. The presence of such dense, target-like distractors inherently challenges and potentially violates the core assumption of \textbf{target sparsity}.

\begin{figure}[htbp]
\centering
\includegraphics[scale=0.6]{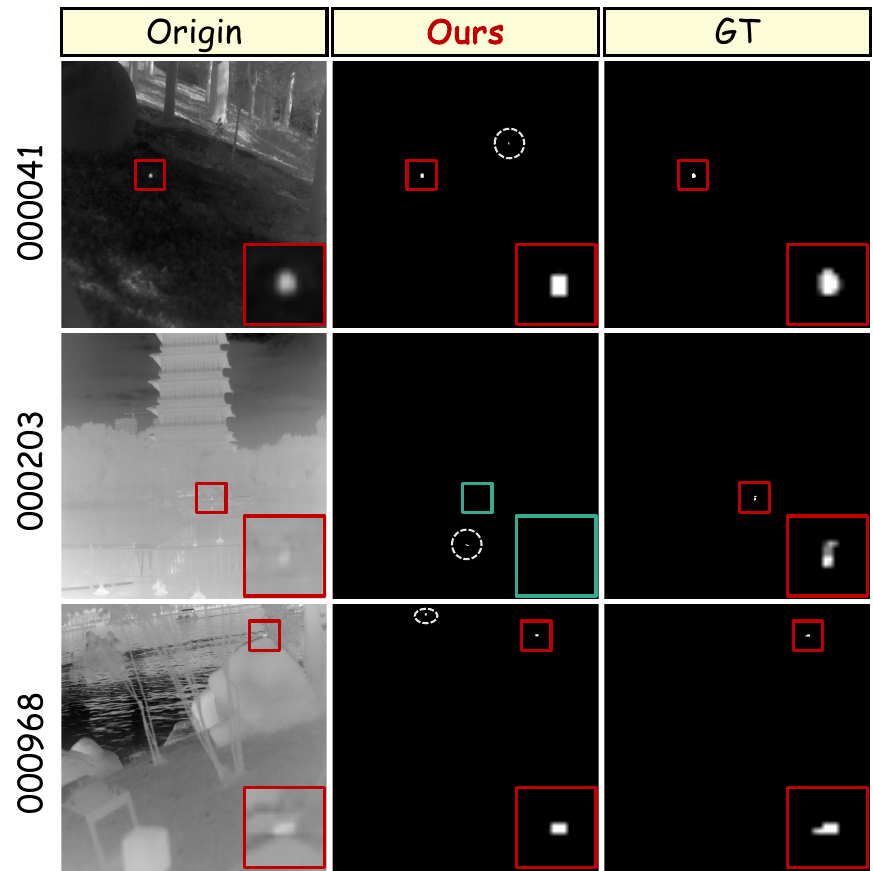}
\caption{Examples of failure cases on the IRSTD-1K dataset. From left to right: original infrared image, DRPCA-Net prediction, and ground truth. Correct detections are marked by red boxes, missed detections by green boxes, and false positives by white circles.}
\label{fig:Failed-image}
\end{figure}

To illustrate these effects more concretely, we provide visual analysis in Fig.~\ref{fig:Failed-image}, which highlights two representative failure patterns encountered by DRPCA-Net in complex terrestrial scenes. First, in Row 2, the model exhibits excessive target suppression, failing to detect a true small target located near the edge of a building. This omission arises from the target’s subtle contrast and proximity to high-frequency textures, causing the model to mistakenly absorb it into the background during the sparse component update. Second, across Rows 1 to 3, background texture misidentification is observed, where false positives (white dotted circles) are generated due to background structures that visually resemble small targets. 

This phenomenon illustrates a potential limitation of leveraging strong model-based priors within deep unfolding architectures. The explicit encoding of the low-rank and sparsity priors provides DRPCA-Net with a powerful inductive bias, enabling it to achieve remarkable detection accuracy and computational efficiency when these assumptions largely hold true, as evidenced by its performance on the first three datasets. However, when faced with data distributions that significantly deviate from these priors, particularly violating the sparsity constraint as observed in parts of IRSTD-1K, the very priors that confer advantages under aligned conditions may impose constraints on the model's flexibility, potentially limiting its performance ceiling relative to methods employing different inductive biases or purely data-driven feature extraction. This underscores a fundamental trade-off: the significant gains in performance, interpretability, and efficiency afforded by model-based priors are most pronounced when the underlying assumptions resonate with the data's statistical properties.

\subsection{Visual Analysis} \label{subsec:visualization}

\begin{figure*}[htbp]
    \centering
    \includegraphics[width=1.0\textwidth]{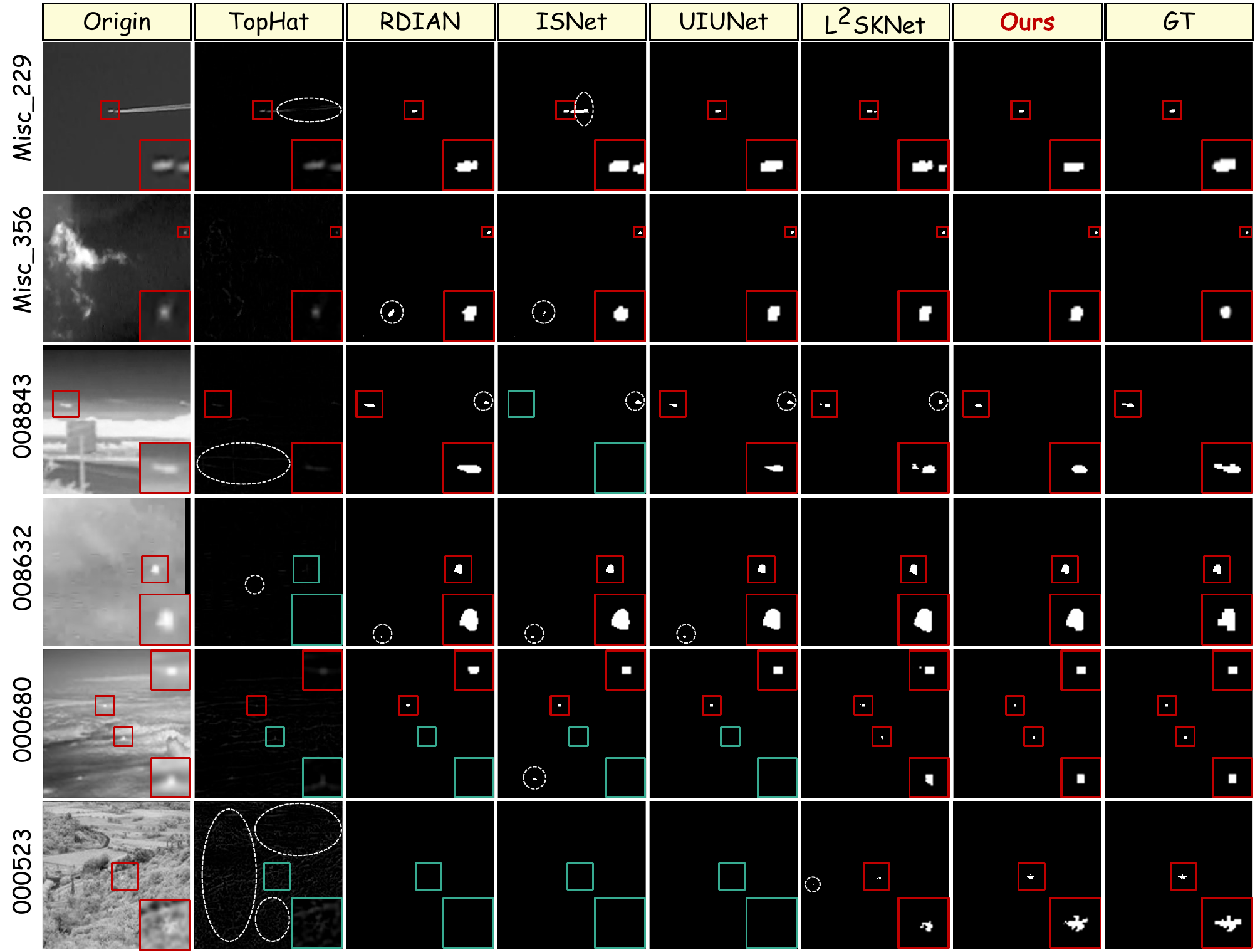}
    \caption{Qualitative comparison on challenging infrared scenes. Columns display the original image, results from TopHat, RDIAN, ISNet, UIUNet, L$^2$SKNet, our DRPCA-Net, and the Ground Truth (GT). Correct detections are marked by red boxes, missed detections by green boxes, and false positives by white dotted circles. Best viewed electronically.}
    \label{fig:visual_comparison}
\end{figure*}

To complement the quantitative evaluations and offer qualitative insights into the operational advantages of DRPCA-Net, we present visualizations of detection results on several challenging exemplars drawn from the SIRSTV1, SIRST-AUG, and NUDT-SIRST datasets. Fig.~\ref{fig:visual_comparison} provides a comparative visual assessment against representative traditional (TopHat) and contemporary deep learning (RDIAN, ISNet, UIUNet, L$^2$SKNet) methods.

\begin{figure*}[htbp]
    \centering
    \includegraphics[width=1.0\textwidth]{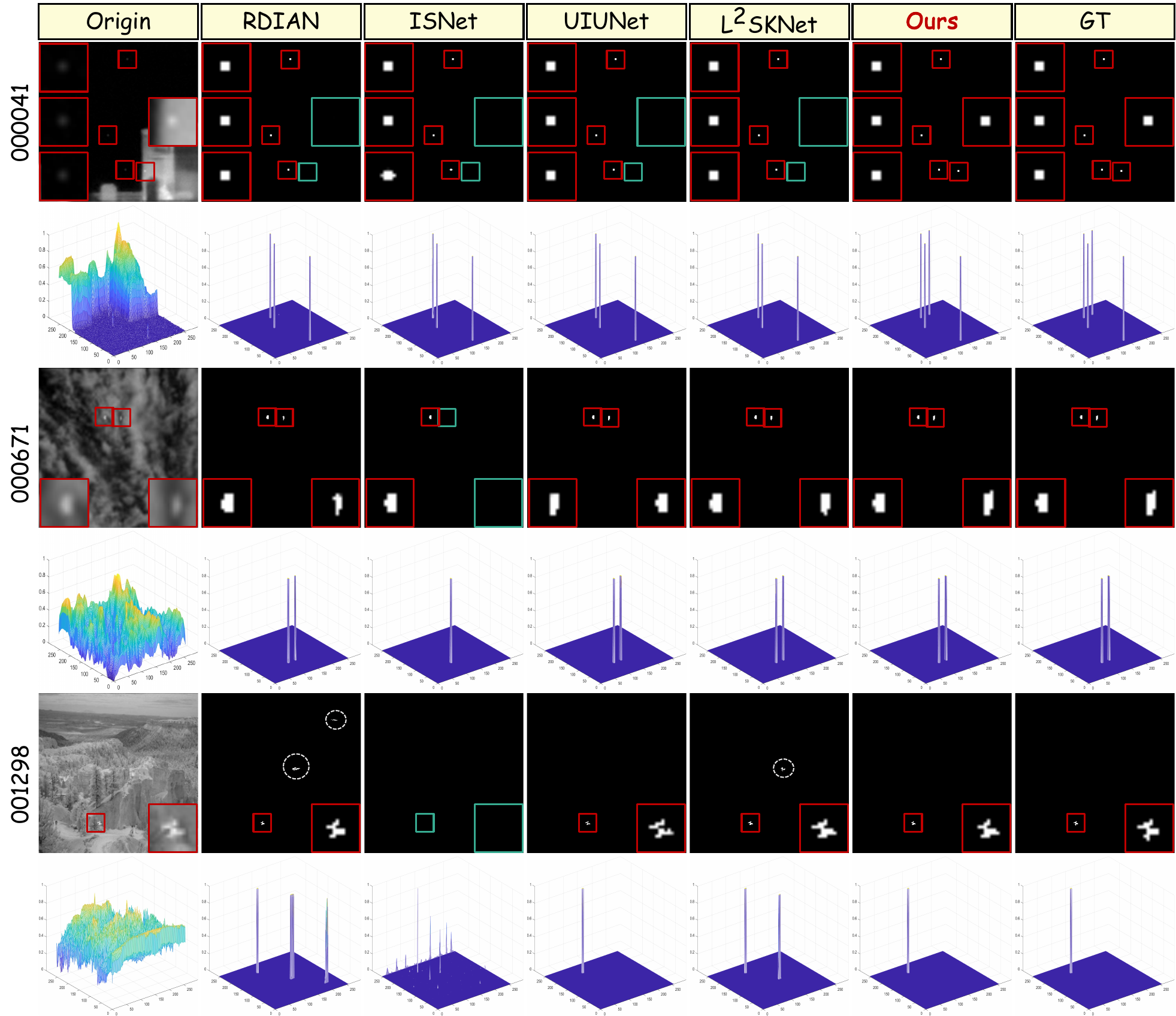}
    \caption{Qualitative comparison on representative challenging infrared scenes featuring multi-target instances, cluttered backgrounds, and target-like distractors. Columns display the original image, results from RDIAN, ISNet, UIUNet, L$^2$SKNet, our DRPCA-Net, and the Ground Truth (GT). Correct detections are marked by red boxes, missed detections by green boxes, and false positives by white dotted circles. Best viewed electronically.}
    \label{fig:visual_comparison2}
\end{figure*}

As visually substantiated in Fig.~\ref{fig:visual_comparison}, traditional methods like TopHat frequently succumb to high false alarm rates and missed detections, particularly in the presence of significant background clutter or low target contrast (e.g., rows 1, 3, 5). While modern deep learning techniques generally exhibit enhanced performance, they remain susceptible to spurious activations induced by complex textures or structured noise, and can still fail to detect targets under challenging conditions (e.g., RDIAN, ISNet in rows 2, 4).

In contrast, DRPCA-Net consistently demonstrates superior robustness and discriminative capability. For instance, in sample \textit{008843} (row 4), while several competing methods (RDIAN, UIUNet, L$^2$SKNet) generate more spatially extensive segmentations, they concurrently produce numerous false positives. DRPCA-Net, however, yields a more precise localization, effectively isolating the true target while adeptly suppressing the surrounding distractors. This highlights its capacity for high-fidelity discrimination, prioritizing detection accuracy over potentially noisy shape completion. Furthermore, in challenging NUDT-SIRST scenes (\textit{000680}, \textit{000523}; rows 5, 6), DRPCA-Net achieves remarkable alignment with the ground truth, successfully identifying all targets without introducing false alarms, thereby showcasing an excellent balance between sensitivity and specificity. Its performance in diverse scenarios, including complex aerial backgrounds (\textit{Misc\_229}, \textit{Misc\_356}; rows 1, 2), further underscores its adaptability.

For a more comprehensive assessment of detection performance, we present a set of diverse visualization examples from the NUDT-SIRST dataset in Fig.~\ref{fig:visual_comparison2}, covering a wide range of challenging scenarios beyond single-target cases. The selected samples include: (1) multi-target scenes with both spatially scattered and closely locatedtargets (row 1 and 3), (2) high-clutter backgrounds such as textured terrain or noisy environments that are prone to false alarms (row 3 and row 5), and (3) target-like distractors that often mislead methods relying on local intensity contrast (row 5). These cases are designed to evaluate the robustness and generalization capability of each method under complex conditions. We also provide 3D surface plots of the predicted soft outputs, which intuitively reveal the response distributions across the spatial domain. Compared with other approaches, our DRPCA-Net consistently produces sharp and sparse peaks corresponding to true targets, while maintaining low and flat responses over background clutter. This demonstrates its superior ability to enhance target saliency and suppress false positives. In contrast, competing methods often exhibit either overly flat responses (missed detections) or noisy activations in the background (false alarms), which are clearly observable from their surface maps.

\begin{figure}[htbp]
\centering
\includegraphics[scale=0.295]{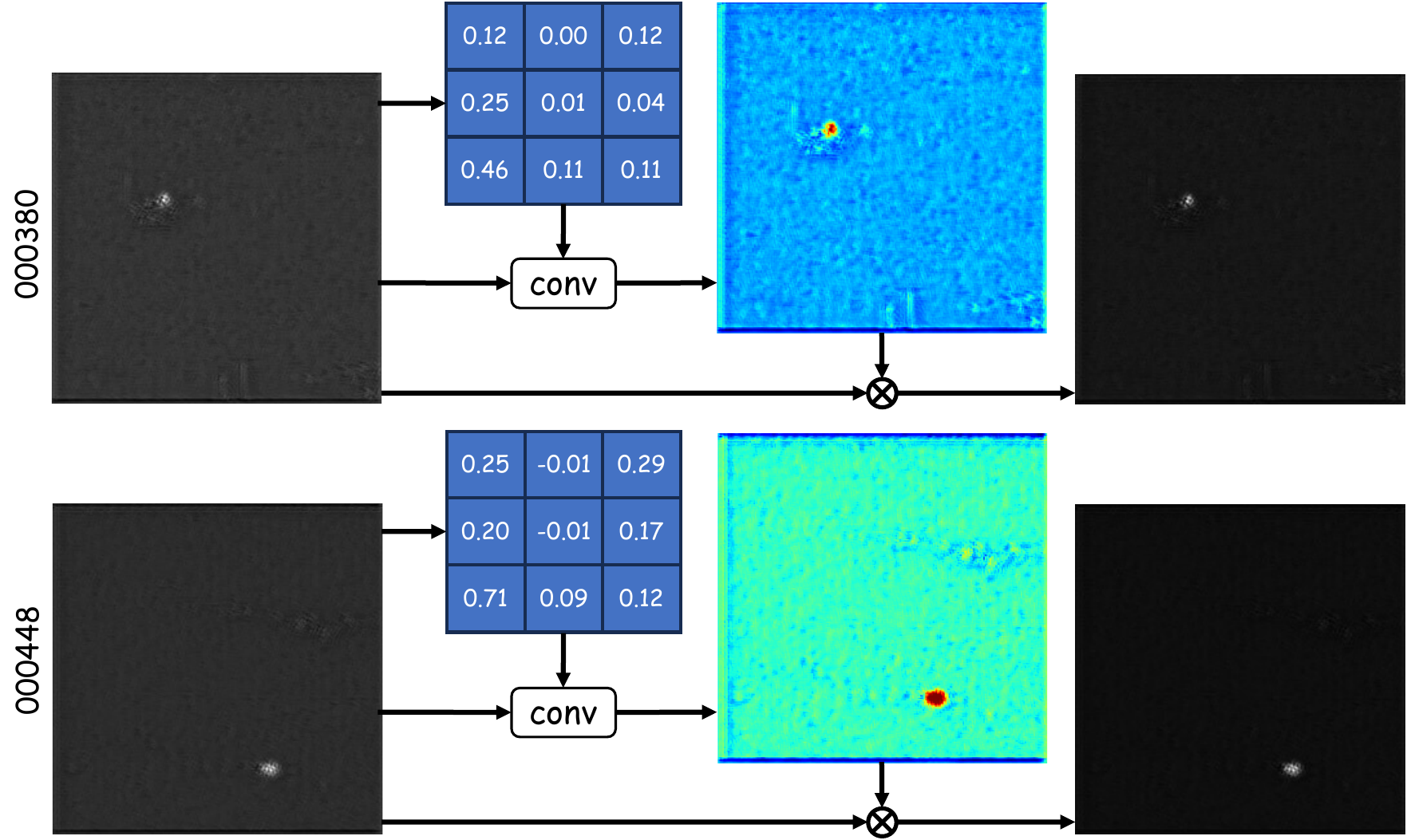}
\caption{Visualization of the DSA module on two infrared target images from the NUDT-SIRST dataset.Red areas in attention maps indicate strong focus on target regions.}
\label{fig:DSA_visualization}
\end{figure}

The effectiveness of the proposed DSA module is further illustrated through visual analysis on representative samples from the NUDT-SIRST dataset, as shown in Fig.~\ref{fig:DSA_visualization}. Each row corresponds to one sample and sequentially displays: the input feature map, the dynamically generated $3\times3$ convolution kernel, the resulting spatial attention map, and the final output after attention modulation. The DSA module generates sample-specific kernels that adapt to the spatial distribution of scene content. The resulting attention maps clearly highlight target regions with strong activations (in red) while suppressing surrounding background clutter. In the final output, target features are selectively enhanced and background interference is effectively attenuated, demonstrating the DSA module’s ability to improve target perception under low signal-to-noise conditions and enhance robustness in complex infrared environments.

To further investigate the scene adaptability of our DRPCA-Net, we conduct a fine-grained evaluation across different background scene types. We categorize the test images from SIRST V1, NUDT-SIRST, and SIRST-Aug datasets into four major scene types based on the semantic context of the target region: sky, ground, building, and ocean. The dataset contains 983 images labeled as sky scenes, 202 as ground scenes, 83 as building scenes, and 27 as ocean scenes.

\begin{figure}[htbp]
\centering
\includegraphics[scale=0.3]{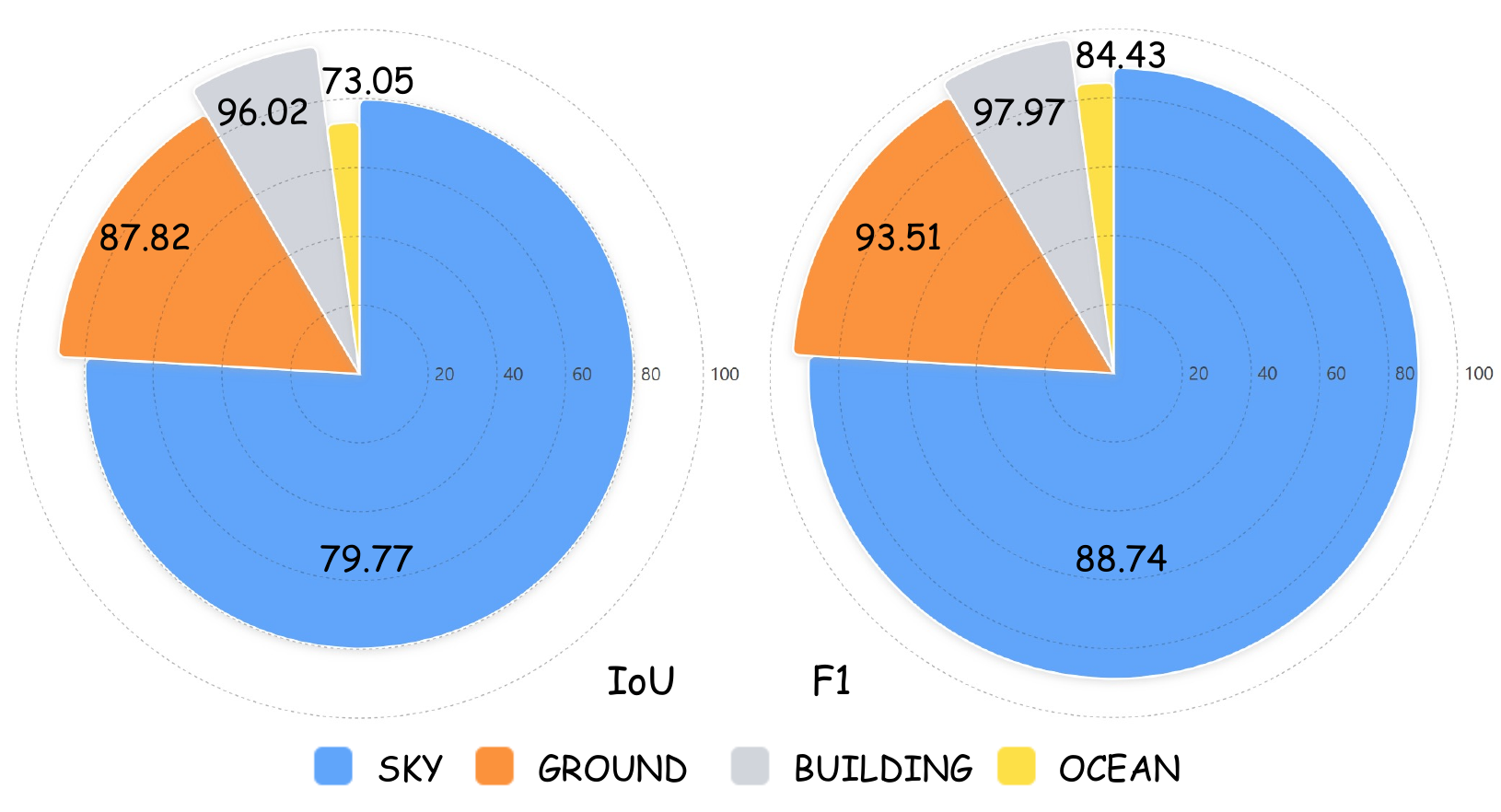}
\caption{Scene-wise performance of DRPCA-Net across four background categories (SKY, GROUND, BUILDING, OCEAN), evaluated by both mean Intersection over Union (IoU, left) and F$_1$ score (right). In each polar chart, the angular width of a sector reflects the proportion of test images belonging to the corresponding scene type, while the radial length indicates the average value of the corresponding metric.}
\label{fig:Miou_in_different_scenesn}
\end{figure}

From the results illustrated in Fig.~\ref{fig:Miou_in_different_scenesn}, we observe that DRPCA-Net performs more robustly in structured scenes, while its performance slightly degrades in homogeneous or low-texture backgrounds. This trend aligns well with the underlying modeling principles of DRPCA-Net. In structured environments, the background often exhibits strong spatial regularity and redundancy, which conforms to the low-rank assumption used in background modeling. Simultaneously, infrared targets in these scenes typically appear as localized anomalies, satisfying the sparse prior applied to the target component. In contrast, unstructured scenes like sky or ocean may violate the low-rank assumption due to stochastic noise and lack of spatial correlation. Targets in such conditions also tend to have lower contrast and less distinct spatial footprints, making them harder to isolate under sparse decomposition. This analysis highlights the strengths and limitations of DRPCA-Net under different scene types and validates the theoretical motivations behind its design.

\section{Conclusion} \label{sec:conclusion}

In this paper, we introduced DRPCA-Net, a novel deep unfolding network architecture that effectively reinvigorates the classical Robust Principal Component Analysis framework for the challenging domain of infrared small target detection. By meticulously unfolding the RPCA optimization process into a dynamic network structure, DRPCA-Net incorporates two principal designs to overcome the limitations of static model assumptions and conventional deep learning approaches. Firstly, a dynamic parameter generation mechanism, realized through a lightweight hypernetwork, imparts input-conditioned adaptability to the unfolding dynamics, enhancing robustness across diverse operational scenarios. Secondly, the DRG module facilitates sophisticated contextual feature refinement and more accurate low-rank background component estimation, thereby significantly improving the discriminability of subtle targets against intricate clutter. 
Extensive experiments demonstrate DRPCA-Net's compelling potential of harmonizing principled model-based priors with the adaptive power of deep learning to advance the state of the art in infrared small target detection.

\section*{Acknowledgment}

The authors would like to thank the editor and the anonymous reviewers for their critical and constructive comments and suggestions.
We acknowledge the Tianjin Key Laboratory of Visual Computing and Intelligent Perception (VCIP) for their essential resources. Computation is partially supported by the Supercomputing Center of Nankai University (NKSC).

\bibliographystyle{IEEEtran}
\bibliography{./reference.bib}


\begin{IEEEbiography}[{
\includegraphics[width=1.45in,height=1.3in,clip,keepaspectratio]{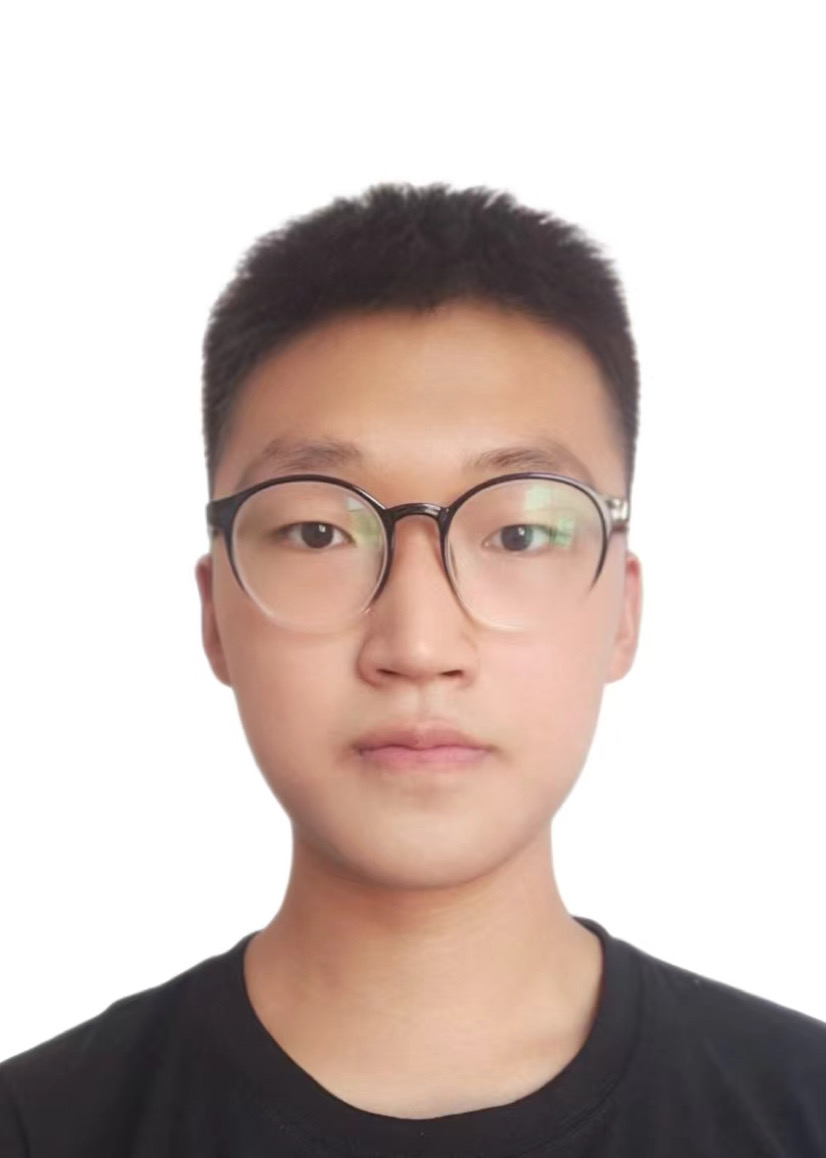}}]{Zihao Xiong} is currently pursuing his Bachelor's degree in Electronic Information Engineering at the School of Information Science and Engineering, Henan University of Technology. He is also a research intern at the PCA Lab, VCIP, College of Computer Science, Nankai University. His research interests include infrared small target detection, tracking, and super-resolution.
\end{IEEEbiography}

\begin{IEEEbiography}[{\includegraphics[width=1in,height=1.25in,clip,keepaspectratio]{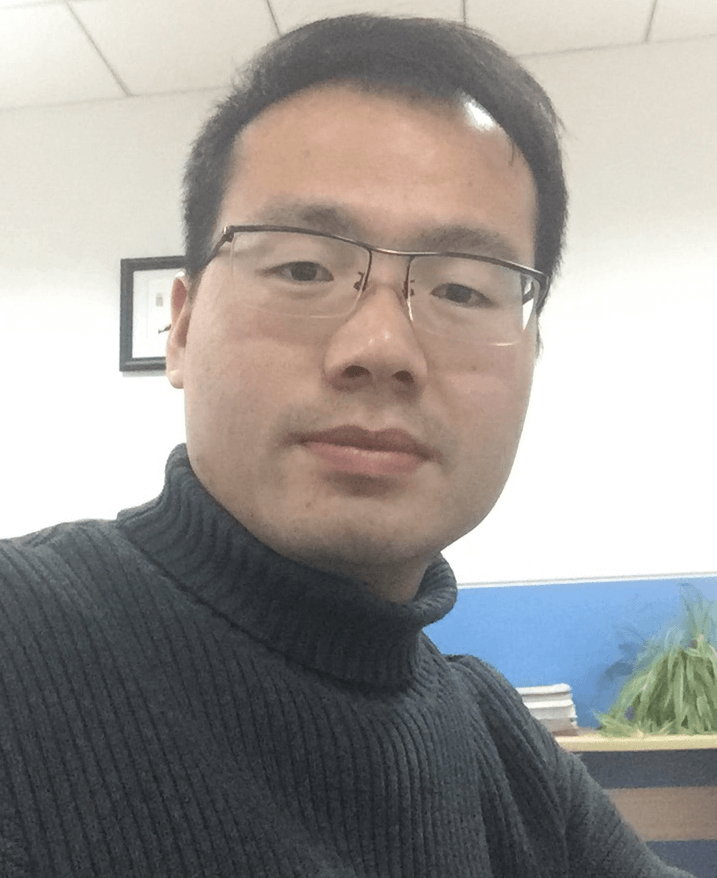}}]{Fei Zhou}
  received his M.S. degree in electronic engineering from Xinjiang University in 2017, and received Ph.D. degree from College of Electronic and Information Engineering, Nanjing University of Aeronautics and Astronautics. He is currently an Associate Professor with College of Information Science and Engineering, Henan University of Technology, Zhengzhou, China. His main research interests are signal processing, target detection and image processing. He has published 10+ papers in journals and conferences, such as TGRS, TAES, WACV, etc.
\end{IEEEbiography}

\begin{IEEEbiography}[{\includegraphics[width=1in,height=1.25in,clip,keepaspectratio]{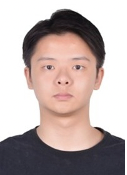}}]{Fengyi Wu}
received his dual B.E. degree in Electric and Electronic Engineering from both University of Electronic Science and Technology of China (UESTC) and University of Glasgow (UoG) in 2021 and is currently chasing a PhD degree at the School of Information and Communication Engineering, UESTC. His current interests include image processing, computer vision, and interpretable target detection.
\end{IEEEbiography}

 \begin{IEEEbiography}[{\includegraphics[width=1in, height=1.4in, clip, keepaspectratio]{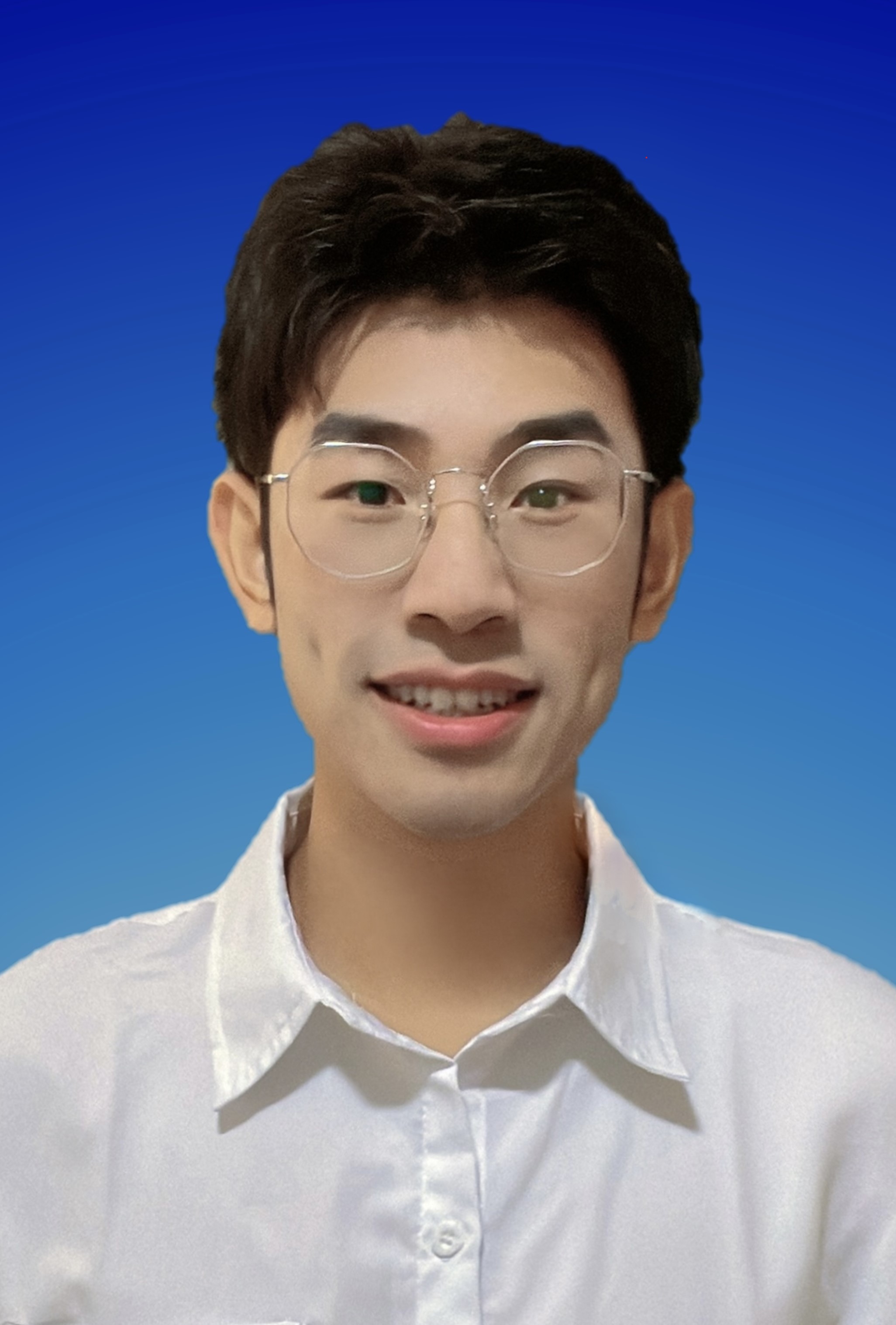}}]{Shuai Yuan}
 received the B.S. degree from Xi'an Technological University, Xi'an, China, in 2019. He is currently pursuing a Ph.D. degree at Xidian University, Xi’an, China. He is currently studying at the University of Melbourne as a visiting student, working closely with Dr. Naveed Akhtar. His research interests include infrared image understanding, remote sensing, and deep learning.
 \end{IEEEbiography}
 
\begin{IEEEbiography}[{\includegraphics[width=1in,height=1.25in,clip,keepaspectratio]{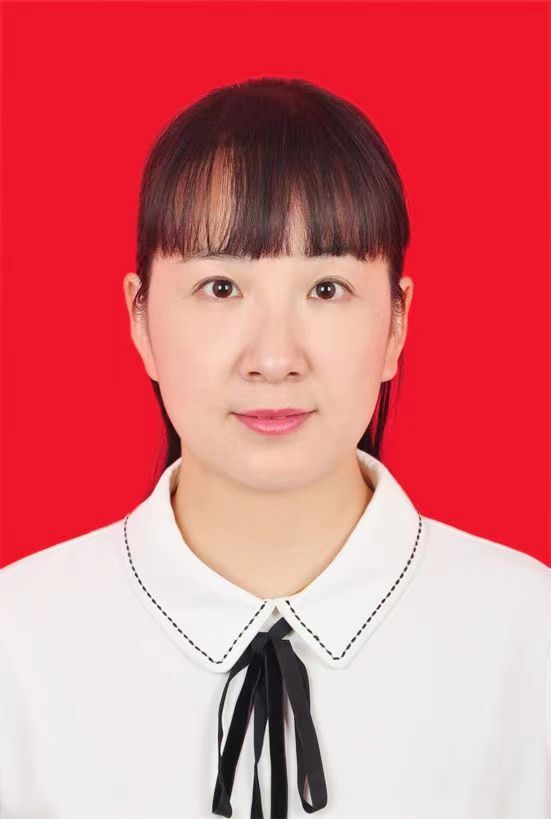}}]{Maixia Fu}
received the B.S. degree in electronic science and technology from Information Engineering University, Zhengzhou, China, in 2003, the M.S. degree in information and communication engineering from Zhengzhou University, Zhengzhou, China, in 2009, and the Ph.D. degree in optical engineering from Capital Normal University, Beijing, China, in 2017. She is currently an Associate Professor and master’s Supervisor with the College of Information Science and Engineering, Henan University of Technology, Zhengzhou, China. Her research interests include signal processing, spectrum analysis, optoelectronic devices designing, and remote sensing.
\end{IEEEbiography}

\begin{IEEEbiography}
[{\includegraphics[width=1in,height=1.25in,clip,keepaspectratio]{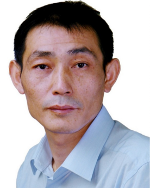}}]{Zhenming Peng}
(Member, IEEE) received a Ph.D. degree in geodetection and information technology from the Chengdu University of Technology, Chengdu, China, in 2001.
From 2001 to 2003, he was a Post-Doctoral Researcher with the Institute of Optics and Electronics, Chinese Academy of Sciences, Chengdu. He is a Professor with the University of Electronic Science and Technology of China, Chengdu. His research interests include image processing, signal processing, and target recognition and tracking.
\end{IEEEbiography}

\begin{IEEEbiography}[{
    \includegraphics[width=1.45in,height=1.3in,clip,keepaspectratio]{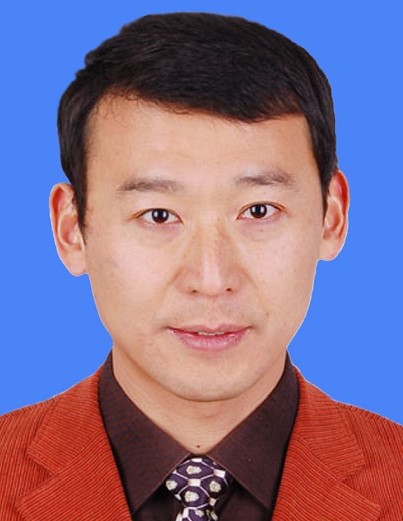}}]{Jian Yang} received the PhD degree from Nanjing University of Science and Technology (NJUST) in 2002, majoring in pattern recognition and intelligence systems. From 2003 to 2007, he was a Postdoctoral Fellow at the University of Zaragoza, Hong Kong Polytechnic University and New Jersey Institute of Technology, respectively. From 2007 to present, he is a professor in the School of Computer Science and Technology of NJUST. Currently, he is also a visiting distinguished professor in the College of Computer Science of Nankai University. His papers have been cited over 50000 times in the Scholar Google. His research interests include pattern recognition and computer vision. Currently, he is/was an associate editor of Pattern Recognition, Pattern Recognition Letters, IEEE Trans. Neural Networks and Learning Systems, and Neurocomputing. He is a Fellow of IAPR. 
\end{IEEEbiography}

\begin{IEEEbiography}[{\includegraphics[width=1in,height=1.25in,clip,keepaspectratio]{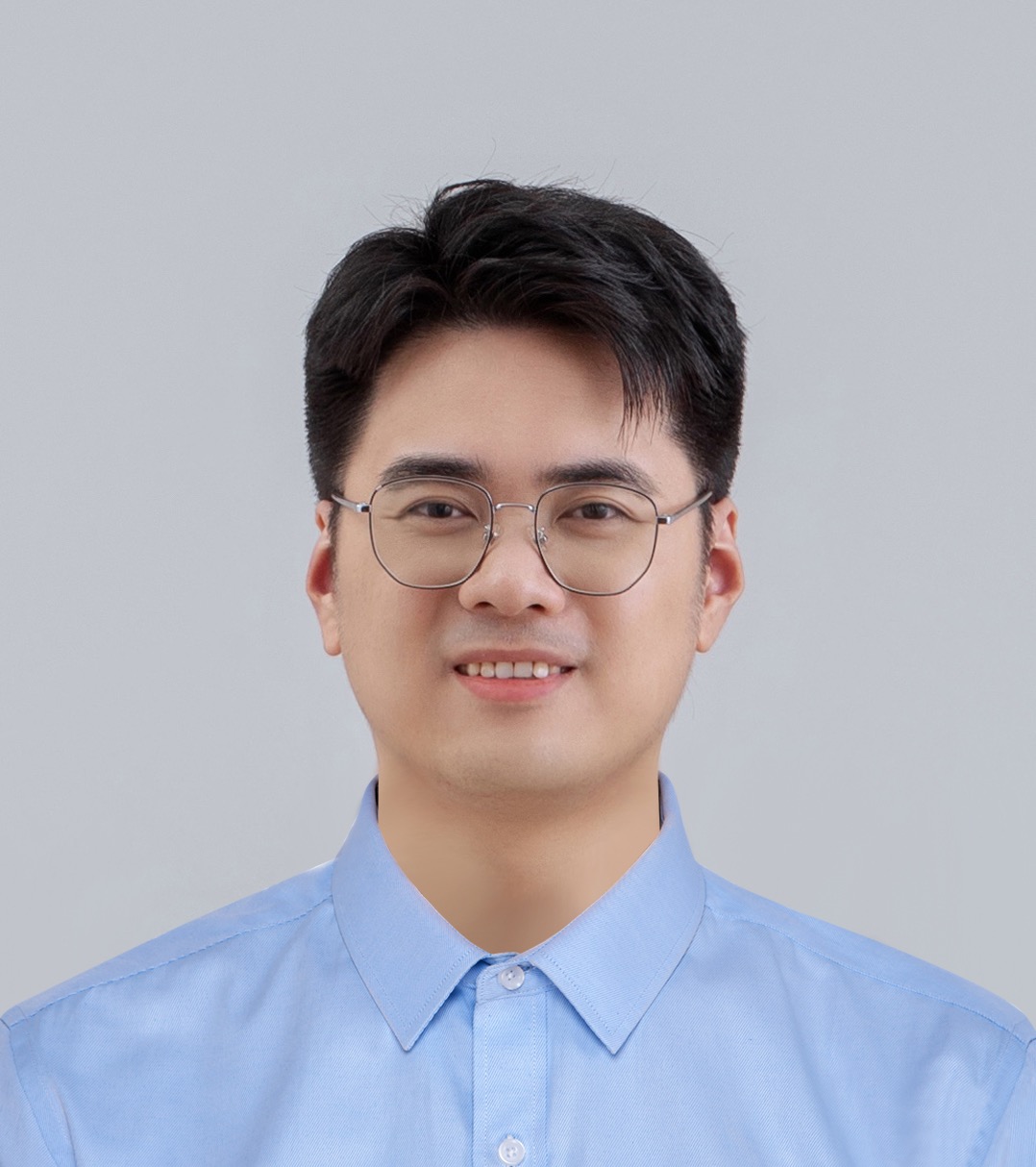}}]{Yimian Dai} (Member, IEEE) received the B.E. degree in information engineering and the Ph.D. degree in signal and information processing from Nanjing University of Aeronautics and Astronautics, Nanjing, China, in 2013 and 2020, respectively.
From 2021 to 2024, he was a Postdoctoral Researcher with the School of Computer Science and Engineering, Nanjing University of Science and Technology, Nanjing, China. 
He is currently an Associate Professor with the College of Computer Science, Nankai University, Tianjin, China.
His research interests include computer vision, deep learning, and their applications in remote sensing.
For more information, please visit the link (\href{https://yimian.grokcv.ai/}{https://yimian.grokcv.ai/}).
\end{IEEEbiography} 

\end{document}